\documentclass[runningheads]{llncs}

\usepackage{eccv}

\usepackage{booktabs}
\usepackage{graphicx}
\usepackage[table]{xcolor}
\usepackage{multirow}
\usepackage{makecell}

\usepackage{xcolor}
\definecolor{darkblue}{RGB}{0,0,250}
\definecolor{darkred}{RGB}{250,0,0}

\usepackage{booktabs}   
\usepackage{tabularx}   
\usepackage{array}      

\usepackage{eccvabbrv}

\usepackage{graphicx}
\usepackage{booktabs}

\usepackage[accsupp]{axessibility}  

\usepackage{hyperref}

\usepackage{orcidlink}

\begin{document}

\title{BrepLLM: Enabling Large Language Models to Understand Boundary Representations} 

\titlerunning{BrepLLM}




\author{
Liyuan Deng\inst{1,2}\textsuperscript{$\dagger$}
\and
Hao Guo\inst{1}\textsuperscript{$\dagger$}
\and
Yongkang Dai\inst{1}
\and
Yunpeng Bai\inst{3}
\and
Yifan Zhu\inst{1}
\and
Yuanyuan Gao\inst{4}
\and
Huaxi Huang\inst{2}\textsuperscript{*}
\and
Yilei Shi\inst{1}\textsuperscript{*}
}

\authorrunning{L.~Deng et al.}

\institute{
Northwestern Polytechnical University\\
\email{\{dly,yilei\_shi\}@mail.nwpu.edu.cn}
\and
Shanghai Artificial Intelligence Laboratory\\
\email{huanghuaxi@pjlab.org.cn}
\and
National University of Singapore
\and
Hong Kong University of Science and Technology
}

\maketitle
\begingroup \renewcommand\thefootnote{} \footnotetext{\textsuperscript{$\dagger$}Equal contribution. \textsuperscript{*}Corresponding author.} \addtocounter{footnote}{-1} \endgroup
\begin{abstract}

Current token-sequence-based Large Language Models (LLMs) struggle to directly process 3D Boundary Representation (B-rep) models that contain complex geometric and topological information. To this end, we propose BrepLLM, the first multimodal framework that enables LLMs to directly parse and reason over raw B-rep data. BrepLLM adopts a two-stage training pipeline: cross-modal alignment pre-training and two-stage LLM fine-tuning. In the first stage, we design an adaptive UV sampling strategy to convert B-reps into graph representations that integrate geometric and topological information. Subsequently, we construct a hierarchical BrepEncoder to extract features from geometric elements (faces and edges) and topology, generating a global token and a sequence of node tokens. Then, via contrastive learning, we conduct an initial alignment between this global token and the text embeddings of a frozen CLIP text encoder (ViT-L/14). In the second stage, we integrate the pre-trained BrepEncoder into the LLM and employ a two-stage progressive strategy to align the sequence of node tokens: (1) training an MLP-based semantic mapping network that utilizes the prior knowledge of a 2D-VLM to align the B-rep representation to the 2D visual semantic space; (2) utilizing LoRA for parameter-efficient fine-tuning of the Q-Former and the LLM backbone network to achieve the final 3D-language generation capability. Furthermore, we construct the Brep2Text dataset, which contains 269,444 B-rep and text question-answer pairs. Experiments demonstrate that BrepLLM achieves SOTA performance on 3D object classification and captioning tasks.The project page is available at \url{https://user-deng.github.io/BrepLLM/}.
\end{abstract}

\section{Introduction}
\label{sec:intro}

Large language models (LLMs) have recently achieved remarkable progress in natural language processing
~\cite{anil2023palm2technicalreport,openai2024gpt4technicalreport,jiang2024mixtralexperts}, computer vision
~\cite{liu2023visualinstructiontuning,li2023blip2bootstrappinglanguageimagepretraining,bai2023qwenvlversatilevisionlanguagemodel},and 3D understanding
~\cite{xu2024pointllmempoweringlargelanguage,zhu2024llava,tang2024minigpt3defficientlyaligning3d},establishing a unified foundation for multimodal intelligence. 
In industrial applications, Computer-Aided Design (CAD) plays an indispensable role, widely used in manufacturing, aerospace, and architecture~\cite{heidari2025geometricdeeplearningcomputeraided}. CAD models are precisely defined by Boundary Representations (B-reps), which serve as the predominant format for expressing complex geometries in freeform surface modeling~\cite{xu2024brepgenbrepgenerativediffusion}. Unlike point clouds or triangle meshes, B-reps  encode exact parametric surfaces together with a watertight topology and explicit topological adjacency, making them both information-rich and structurally constrained for downstream geometric reasoning and symbolic manipulation.
Enabling LLMs to directly parse and reason over native B-rep models would drive significant advances in industrial model design and intelligent manufacturing.


However, existing large model frameworks still cannot directly process the native geometric and topological structures of CAD models. Instead of directly modeling B-reps, they typically convert CAD into executable intermediate representations—such as command sequences, sketch operation sequences, or program code—that are later executed by a CAD kernel. For example, CAD-MLLM~\cite{Cad-mllm} generates CAD command sequences to synthesize parametric models from text, images, or point clouds. CadVLM and CAD-LLM~\cite{cadvlm,Cad-llm} leverage pretrained models to generate sketch sequences for tasks such as sketch auto-completion and constraint reasoning, while cadrille and CAD-Coder~\cite{doris2025cadcoderopensourcevisionlanguagemodel,kolodiazhnyi2026cadrillemultimodalcadreconstruction} use LLMs to generate CAD programs for indirect 3D construction. Although these methods have shown progress for generation and editing, their operation space is largely procedural rather than the underlying B-rep entities (faces/edges) and topology, which limits genuine geometric or topological reasoning. Moreover, construction sequence data is extremely scarce, whereas native B-rep models are an industry-standard format that is abundant and easier to acquire at scale.

\begin{figure}[t!]
  \centering
  \includegraphics[width=\columnwidth]{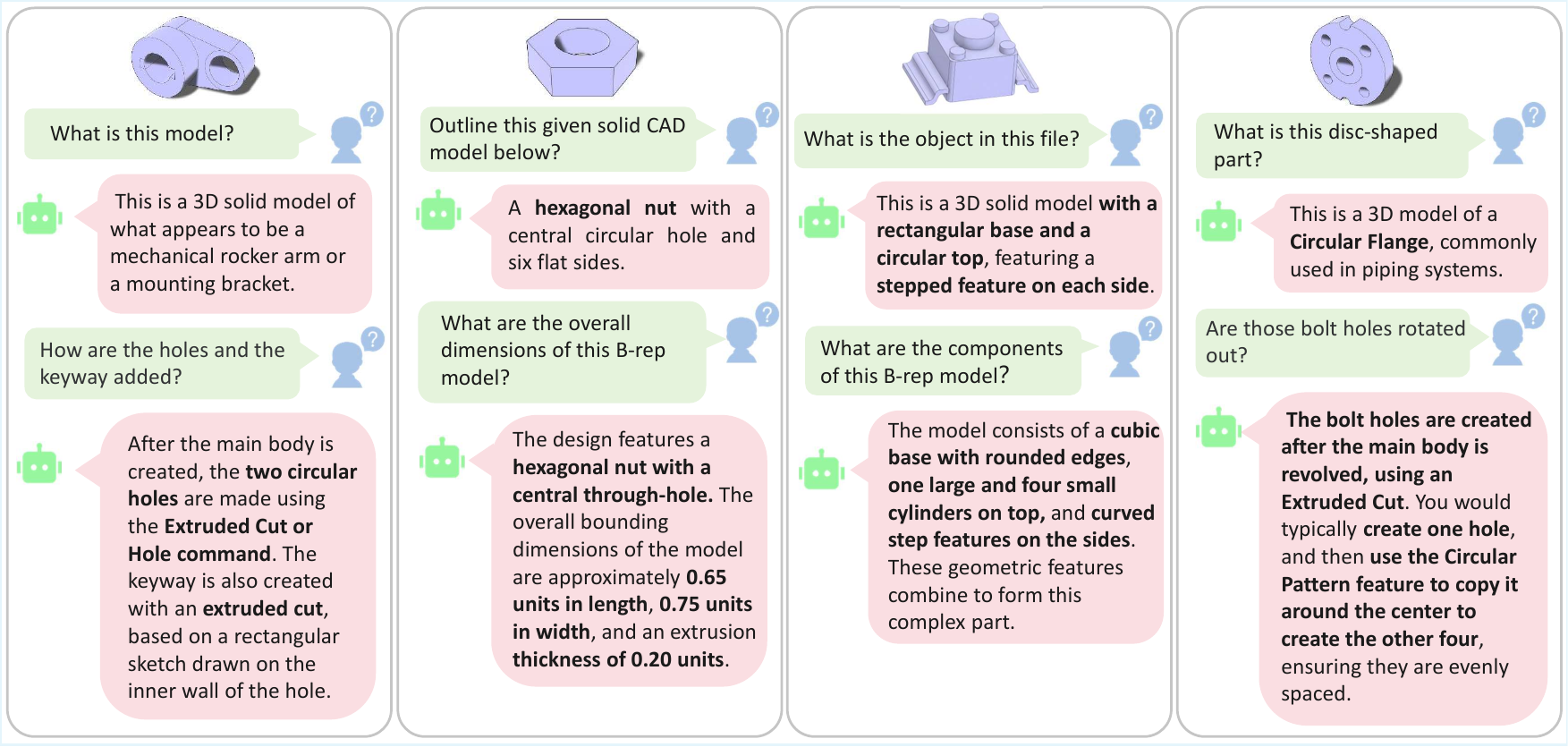}
  \caption{We introduce BrepLLM, a multi-modal large language model capable of understanding B-rep models. By natively perceiving the topology and geometry of B-reps, it enables multi-turn interactions without occlusion or viewpoint dependency.}
  \label{fig:brepllm_interaction}
  \vspace{-0.70cm}
\end{figure}

To bridge this gap, we propose \textbf{BrepLLM}, the first large language model framework for directly understanding B-rep (Figure~\ref{fig:brepllm_interaction}). BrepLLM employs a two-stage training pipeline, consisting of \textbf{Cross-modal Alignment Pre-training} and \textbf{Two-stage LLM Fine-tuning} (Figure~\ref{fig:framework}). In the cross-modal alignment stage, we introduce an adaptive UV sampling strategy that converts B-reps  into graph representation. We design BrepEncoder, a hierarchical architecture that jointly extracts local geometric features of faces, edges, and topological features, which are then globally aggregated to produce a representation. We then employ a CLIP-style contrastive learning strategy~\cite{clip} to align the global representation produced by BrepEncoder with the frozen CLIP text encoder (ViT-L/14)~\cite{radford2021learningtransferablevisualmodels}, thereby constructing a unified geometry–language embedding space.

In the LLM fine-tuning stage, we integrate the pretrained BrepEncoder into a large language model and adopt a two-stage progressive alignment strategy. We first leverage the alignment priors of 2D vision–language models (2D-VLMs) to establish an initial correspondence between B-reps and text. We then perform LoRA-based fine-tuning to incorporate B-rep representations into the LLM’s semantic space. Furthermore, we construct the \textbf{Brep2Text} dataset, comprising 269,444 high-quality Brep–language pairs. This dataset serves as a large-scale benchmark for training and evaluating models on B-rep-centric language understanding and reasoning tasks.


Our key contributions are as follows:
\begin{itemize}
  \item We propose BrepLLM, the first framework that enables large language models to directly parse and reason over native B-rep models.
  \item We introduce an adaptive UV sampling strategy and a hierarchical BrepEncoder to model geometry and topology, producing multi-granularity representations over faces, edges, and their connectivity.
  \item 
  We adopt a progressive alignment strategy that transfers alignment priors from 2D-VLMs and gradually integrates B-rep geometric features into the semantic space of large language models via LoRA-based fine-tuning.
  
  \item We establish the first large-scale benchmark for B-rep-centric language understanding tasks, comprising 269,444 high-quality B-rep models paired with natural language question–answer examples. 
\end{itemize}

\section{Related Work}
\label{sec:formatting}

\subsection{CAD in LLMs}

Recent work on integrating LLMs into CAD has predominantly followed a procedural representation paradigm, converting parametric CAD into command sequences or domain-specific languages (DSLs) amenable to generation and editing. In the realm of sequence generation, CAD-LLM~\cite{Cad-llm} and CadVLM~\cite{cadvlm} tackle sketch auto-completion, while Text2CAD~\cite{text2cad}, Text to CADQuery~\cite{xie2025texttocadquerynewparadigmcad}, and Query2CAD~\cite{query2cad} pioneer text-conditioned CAD command sequence synthesis paradigm. Building on these foundations, CAD-LLama~\cite{cadllama}, CAD-coder~\cite{doris2025cadcoderopensourcevisionlanguagemodel}, LLM4CAD~\cite{llm4cad}, and FlexCAD~\cite{flexcad} further enhance code construction capabilities via instruction tuning. Multimodal and interactive applications have likewise flourished: CAD-MLLM~\cite{Cad-mllm} and CAD-GPT~\cite{Cad-gpt} supported multimodal sequence generation, whereas CAD-Editor~\cite{CAD-Editor}, CAD Translator~\cite{cadtranslator}, BlenderLLM~\cite{blenderllm}, and CAD-Assistant~\cite{Cad-assistant} explore local editing, code migration, and interactive assistance. Nevertheless, these methods primarily operate over procedural sequences rather than directly on the native B-rep entities --- faces, edges, and topological structures --- and thus fall short of direct geometric perception in 3D space. Furthermore, they depend on complete construction history data, which is difficult to obtain: the DeepCAD dataset~\cite{Wu2023PointTV}, which includes build sequences, is only one-fifth the size of the ABC dataset~\cite{abc}, which provides only the final B-rep models. These constraints in data scale and modeling flexibility pose significant bottlenecks when scaling toward larger CAD foundation models.

\subsection{Direct B-rep Learning}

As the standard data format in industrial CAD, B-rep has attracted growing attention for 3D model classification, retrieval, feature recognition, and generation. Especially in the field of 3D generation, learning methods based directly on native B-rep are gradually becoming a research hotspot because they do not rely on construction history or predefined modeling operations. SolidGen~\cite{solidgen} decomposes B-rep generation into joint prediction of vertices, edges, and faces, though it does not support freeform surface and curve types. To overcome these geometric expressiveness limitations, BrepGen~\cite{xu2024brepgenbrepgenerativediffusion} introduces a multi-stage diffusion model for B-rep generation in latent space, and NeuroNURBS~\cite{neuronurbs} employs B-spline surface primitives in place of traditional UV grids for more compact geometric encoding. To improve the structural validity of generated outputs, DTGBrepGen~\cite{dtgbrepgen} proposes disentangled generation of topology and geometry. Additionally, HoLa~\cite{hola} constructs a holistic variational autoencoder that maps complete B-rep models into a unified latent space, substantially simplifying latent diffusion model training. 
Most recently, AutoBrep~\cite{autobrep} adopts an autoregressive architecture for B-rep sequence generation, achieving notable improvements in both generation quality and computational efficiency. In a related direction, BrepARG~\cite{li2026autoregressivegenerationbrepholistic} represents B-rep models as holistic token sequences that jointly encode geometric and topological information, further demonstrating the potential of sequence modeling for native B-rep generation. Despite these advances in native B-rep representation and generation, existing methods mainly focus on reconstruction or generative modeling, and no existing work has yet realized a unified multimodal large language model capable of natively parsing B-rep data and supporting complex engineering reasoning.

\subsection{3D Large Models for Object Understanding}

Recent efforts to integrate LLMs with 3D data have made substantial progress on object-level understanding tasks. Early approaches~\cite{hong20233dllminjecting3dworld} rely on multi-view 2D renderings to leverage existing 2D vision-language models, but such methods are susceptible to spatial occlusion and lose fine-grained 3D geometric detail. The research paradigm has since shifted toward direct processing of 3D native data. Representative works including Point-Bind~\cite{guo2023pointbindpointllmaligning}, PointLLM~\cite{pointllm}, and ShapeLLM~\cite{shapellm} introduce pretrained 3D encoders to achieve end-to-end alignment between point cloud features and the representation space of language models, effectively bridging geometric and semantic information. Building on these, MiniGPT-3D~\cite{tang2024minigpt3defficientlyaligning3d} incorporates 2D priors and proposes a Mixture-of-Query-Experts (MQE) module with a cascaded training strategy, further improving the efficiency and fidelity of mapping point cloud features to the text space. GreenPLM~\cite{greenplm} enables direct transfer of monolingual pretrained language models to other languages via bilingual lexicons. Beyond object-level understanding, recent studies have also extended 3D language-conditioned intelligence toward physically grounded world modeling and embodied action. Mirage2Matter~\cite{gao2026mirage2matterphysicallygroundedgaussian} constructs a physically grounded Gaussian world model from video for scalable embodied training data generation, while HUGE-Bench~\cite{guo2026huge} introduces a benchmark for high-level UAV vision-language-action tasks in 3D digital-twin environments. These efforts highlight the growing importance of connecting 3D perception, language, and action. Despite these advances, existing 3D large multimodal models (3D LMMs) still predominantly operate on discrete point clouds, polygonal meshes, 2D renderings, or scene-level Gaussian representations. Such representations inherently struggle to capture the continuous parametric surfaces and strict topological constraints in native CAD B-rep, leaving precise geometric understanding and reasoning in engineering scenarios an open challenge.

\begin{figure*}[t]
  \centering
  \includegraphics[width=\linewidth]{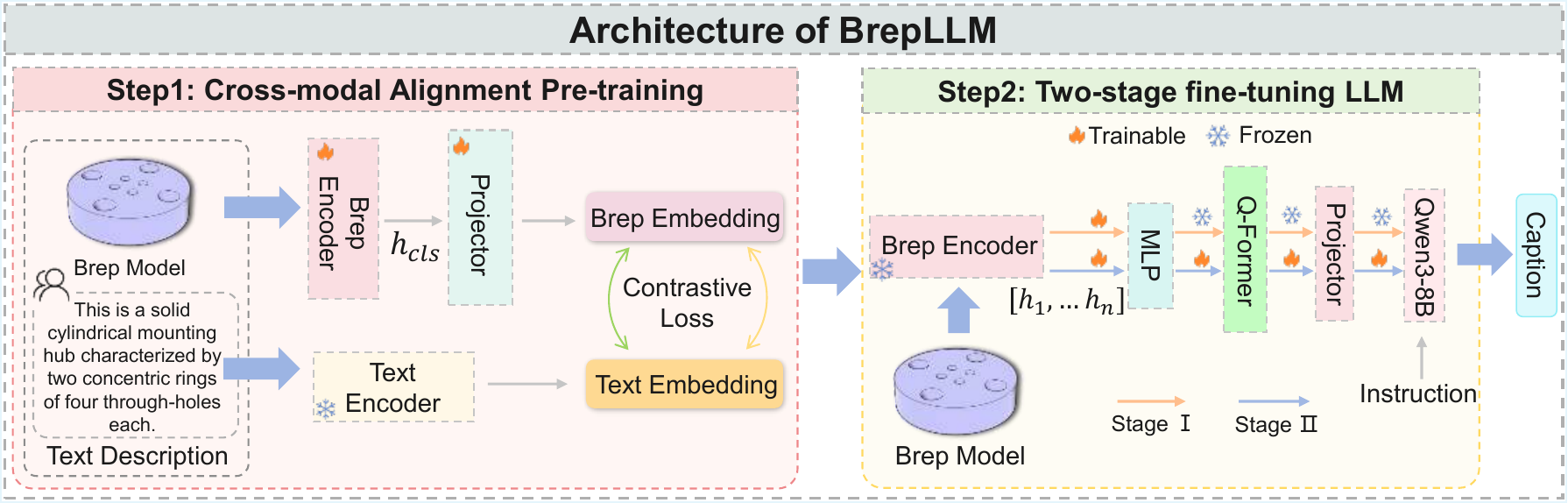}
  \caption{Overview of the BrepLLM architecture. The framework consists of two steps. Step 1 (Left): Cross-modal Alignment Pre-training. BrepEncoder processes the B-rep model to produce a global feature. This feature is then aligned with text embeddings from a frozen CLIP Text Encoder (ViT-L/14) using a contrastive loss.
  Step 2 (Right): Two-stage LLM Fine-tuning. The frozen BrepEncoder's node tokens are progressively aligned with LLM. Stage I trains an MLP to map the node tokens; Stage II fine-tunes the Q-Former and LLM using LoRA.}
  \label{fig:framework}
\end{figure*}

\section{Method}
\subsection{Architecture Overview}
BrepLLM adopts a two-stage training framework: \textbf{Cross-modal Alignment Pre-training} and \textbf{Two-stage LLM Fine-tuning}, as illustrated in Figure~\ref{fig:framework}.

\textbf{Cross-modal Alignment Pre-training} aims to extract high-dimensional geometric and topological features from raw CAD data and align them with natural language descriptions, consisting of three steps.  
\textbf{(1) Graph Representation}: B-rep models are converted into graphs by representing face-adjacency topology as edge, and representing parameterized fine-grained geometry as node via adaptive UV sampling;
\textbf{(2) Feature Encoding}: A hierarchical BrepEncoder processes the graph and extracts both a global token representing the entire B-rep and a sequence of node tokens corresponding to each face;
\textbf{(3) Cross-modal Alignment}: The global B-rep feature is aligned with text embeddings from a frozen CLIP text encoder (ViT-L/14) via CLIP contrastive loss.

\textbf{Two-stage LLM Fine-tuning} freezes the BrepEncoder and fine-tunes the pretrained language model Qwen3-8B~\cite{qwen3technicalreport} in two stages.  
\textbf{Stage I (Geometry-to-Vision Bridging)}: A lightweight MLP is trained to project B-rep features into the Q-former input space;  
\textbf{Stage II (3D–Language Alignment)}: The Q-former and selected LLM layers are fine-tuned using LoRA.


\subsection{Cross-modal Alignment Pre-training}
\begin{figure*}[t]
  \centering
  \includegraphics[width=\linewidth]{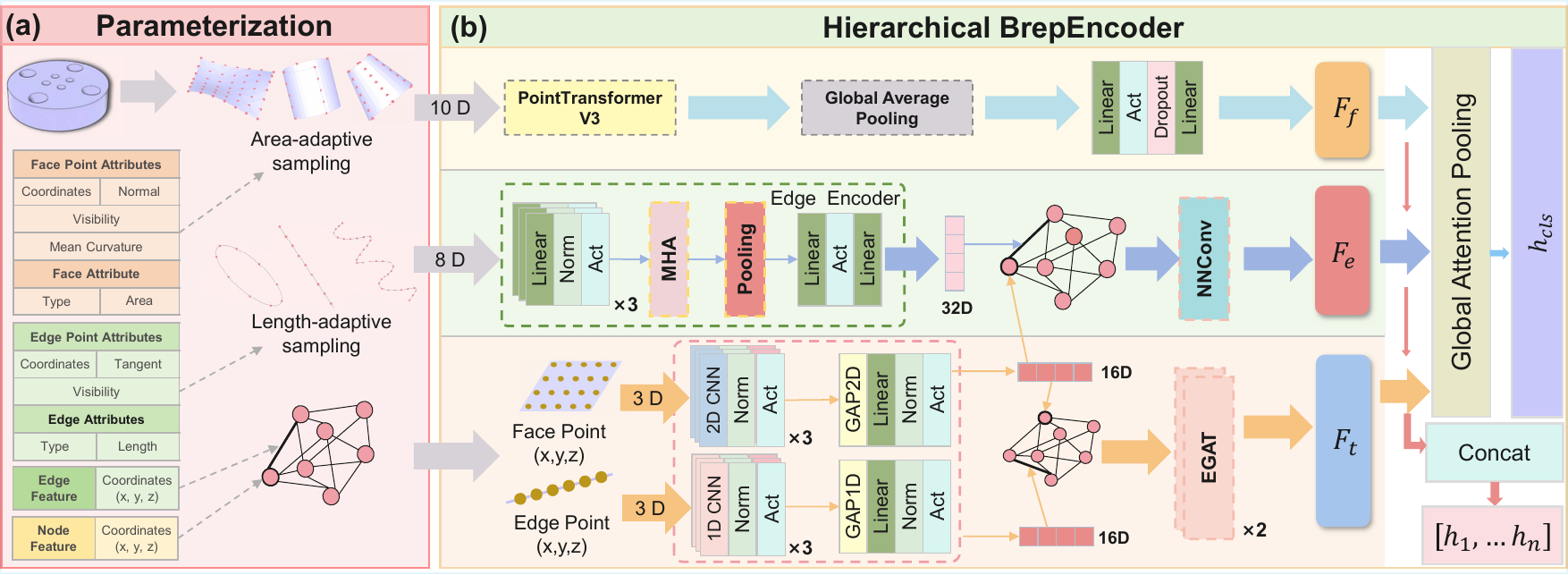}
    \caption{ The overview of BrepEncoder.
    (a) B-rep parameterization using area-adaptive UV sampling for faces and length-adaptive sampling for edges, producing the face attribute tensor $\mathbf{X}_{\mathcal{S}}$ and edge attribution tensor $\mathbf{X}_{\mathcal{C}}$.
    (b) Hierarchical BrepEncoder. Face features $F_f$, edge-conditioned features $F_e$, and global topology features $F_t$ are extracted from per-node tokens $\mathbf{h}_i$. A global graph feature $\mathbf{h}_{\text{cls}}$ is obtained via global attention pooling.
    }
    \vspace{-0.5cm}
  \label{fig:brep_encoder}
\end{figure*}

\noindent\textbf{B-rep Parameterization.} To ensure compatibility with neural network inputs while preserving geometric and topological information, we propose a multi-scale adaptive UV sampling strategy. See Figure~\ref{fig:brep_encoder}(a). This approach constructs a face-edge topological graph structure for joint discretization of faces and edges, where nodes represent parameterized face patches and edges encode adjacency relationships between faces. Face domains employ area-driven grid sampling, while edges adopt length-adaptive point sampling to establish multi-scale geometric representations across local details and global structures.

For face discretization, in the adaptive UV sampling, with given parameter domain $\Omega_{\mathcal{S}} = [u_{\min}, u_{\max}] \times [v_{\min}, v_{\max}]$, where $u_{\min}, u_{\max}$ and $ v_{\min}, v_{\max}$ refer to the two directions minimum and maximum dimensions discretization respectively. The density $N_{\mathcal{S}}$ is computed as:
\vspace{-0.05cm}
\begin{equation}
N_{\mathcal{S}} = N_{\min}^{\mathrm{face}} + \frac{A_{\mathcal{S}} - A_{\min}}{A_{\max} - A_{\min}} \cdot (N_{\max}^{\mathrm{face}} - N_{\min}^{\mathrm{face}})
\end{equation}
\vspace{-0.05cm}
where $A_{\min}/A_{\max}$ denote the minimum/maximum face areas in the model. Each sample point $(u_k, v_l)\in\Omega_{\mathcal{S}}$ yields a 10-dimensional feature tensor $\mathbf{X}_{\mathcal{S}} \in \mathbb{R}^{N_{\mathcal{S}} \times 10}$ by extracting 3D coordinates $\mathbf{P} \in \mathbb{R}^3$, unit normals $\mathbf{n} \in \mathbb{S}^2$, mean curvature $H \in \mathbb{R}$, visibility mask $\mathbf{V} \in \{0,1\}$, face type $\mathbf{t} \in \mathbb{Z}$, and normalized area $a = A_{\mathcal{S}}/A_{\max}$.

For edges, the sampling count $M_{\mathcal{C}}$ is determined by:
\begin{equation}
M_{\mathcal{C}} = M_{\min}^{\mathrm{edge}} + \frac{\ell_{\mathcal{C}} - \ell_{\min}}{\ell_{\max} - \ell_{\min}} \cdot (M_{\max}^{\mathrm{edge}} - M_{\min}^{\mathrm{edge}})
\end{equation}
with $\ell_{\min}/\ell_{\max}$ representing edge length range. Each sample point produces an 8-dimensional sequence feature $\mathbf{X}_{\mathcal{C}} \in \mathbb{R}^{M_{\mathcal{C}} \times 8}$ containing 3D coordinates $\mathbf{Q} \in \mathbb{R}^3$, unit tangents $\boldsymbol{\tau} \in \mathbb{S}^2$, edge type $\mathbf{c} \in \mathbb{Z}$, and normalized length $b = \ell_{\mathcal{C}}/\ell_{\max}$.

This framework provides high-quality input for cross-modal alignment pretraining through $C_{\mathrm{face}} = 10$ node features and $C_{\mathrm{edge}} = 8$ edge features, with face sampling count $N_{\mathcal{S}}$ and edge sampling count $M_{\mathcal{C}}$ each constrained to the range $[32, 64]$, balancing geometric precision and topological integrity.

\noindent\textbf{Hierarchical BrepEncoder}. We design a novel Hierarchical BrepEncoder to encode geometry and topology of a B-rep model. The encoder produces a global graph feature $\mathbf{h}_{\text{cls}}$ and a sequence of node tokens obtained by concatenating three level-wise features. The global feature is used for contrastive learning, and the node tokens are used to fine-tune the LLM. It explicitly decomposes B-rep features into three disentangled, fine-to-coarse hierarchical representations: 1) the fine-grained face feature $F_f$, 2) edge-conditioned feature $F_e$, and 3) global topology feature $F_t$. See Figure~\ref{fig:brep_encoder}(b).

We compute the face feature $F_f$ for geometric details within each face. For this branch, we process the full attribute tensor of each face that contains geometric attributes (e.g., coordinates, normals, mean curvature, etc.). This tensor is fed into a PointTransformerV3 module~\cite{Wu2023PointTV}, which applies self-attention over these sampled attribute points to produce a 32-dimensional feature vector $F_f\in\mathbb{R}^{32}$.

Then we compute the edge-conditioned feature $F_e$ to encode local boundary and neighborhood information defined by adjacent edges. We introduce an Edge Encoder that transforms five attributes of each edge (e.g., coordinates, tangents, edge type, normalized length) into a feature representation capturing edge-level geometry.
This representation is used as the kernel in an NNConv layer, enabling neighbor-to-center face message passing conditioned on the geometry of the shared boundary, resulting in a 32-dimensional feature $F_e\in\mathbb{R}^{32}$.

Finally, we extract the global topological feature $F_t$ of the B-rep model. We first encode the 3D coordinates of faces and edges, using 2D and 1D CNNs, respectively. The resulting features are fed into two layers of EGATConv. Through multi-hop attention propagation, information about the global topology is integrated into each face node, resulting in a 64-dimensional $F_t\in\mathbb{R}^{64}$.

These three features are concatenated to face node $i$:
\begin{equation}
h_i = [ F_{t}^{(i)} \Vert F_{e}^{(i)} \Vert F_{f}^{(i)} ] \in \mathbb{R}^{128}
\end{equation}

For contrastive learning and LLM fine-tuning, the BrepEncoder outputs a global graph feature $\mathbf{h}_{\text{cls}}$ and a sequence of node tokens $[ h_1, h_2, \dots, h_n ]$. The global feature $\mathbf{h}_{\text{cls}}$ serves as a compact representation of the entire B-rep model for contrastive pretraining, while the node tokens $h_i$ encode hierarchical geometric and topological information and are used as inputs for aligning the LLM.



\noindent\textbf{Brep--Text Embedding Alignment.} Directly train a large language model with a complex, hierarchical graph representations as input is challenging. To bridge this gap, we first align the B-rep features with natural language descriptions through contrastive pretraining, mapping both modalities into a shared semantic embedding space. Then, we construct the Brep2Text dataset, which contains a large number of CAD models paired with corresponding textual descriptions. Based on this dataset, we adopt a CLIP-style dual-tower architecture for pretraining. The text tower is a frozen CLIP text encoder (ViT-L/14) \cite{clip}, while the geometry tower comprises our BrepEncoder combined with a projector layer, which is jointly trained to align geometric representations with the text embedding space.

We take the global graph feature $\mathbf{h}_{\text{cls}}$ and feed it into a projector layer. This projects the feature to match the embedding dimension $D$ of the text encoder, resulting in the final B-rep global embedding $\mathbf{z}_{\text{brep}} \in \mathbb{R}^D$. For the text modality, the descriptive caption is fed into the frozen CLIP text encoder, and the output is used as the semantic text representation $\mathbf{z}_{\mathrm{text}}\in\mathbb{R}^{D}$.
We employ a symmetric contrastive loss for training. Given a batch of $N$ matched shape-text pairs, we first $L_2$-normalize their respective embeddings: 
\begin{equation}
\hat{\mathbf{z}}_{\text{brep}} = \mathbf{z}_{\text{brep}} / \Vert\mathbf{z}_{\text{brep}}\Vert_2, \quad
\hat{\mathbf{z}}_{\text{text}} = \mathbf{z}_{\text{text}} / \Vert\mathbf{z}_{\text{text}}\Vert_2
\end{equation}

We then compute a pairwise cosine similarity matrix $S \in \mathbb{R}^{N \times N}$, where $S_{ij} = \hat{\mathbf{z}}_{\text{brep},i} \cdot \hat{\mathbf{z}}_{\text{text},j}$. These similarity scores are scaled by a learnable temperature parameter $\tau$ within the softmax normalization to obtain the shape-to-text distribution $P$ and the text-to-shape distribution $Q$:
\begin{equation}
P_{ij} = \frac{\exp(S_{ij}/\tau)}{\sum_{k=1}^{N} \exp(S_{ik}/\tau)}
\end{equation}
\begin{equation}
Q_{ij} = \frac{\exp(S_{ij}/\tau)}{\sum_{k=1}^{N} \exp(S_{kj}/\tau)}
\end{equation}

Finally, we adopt the symmetric InfoNCE loss, which maximizes the similarity of the $N$ correctly matched pairs (i.e., the diagonal elements of the probability distributions):
\begin{equation}
\mathcal{L}_{\text{CLIP}} = -\frac{1}{2N} \sum_{i=1}^{N} \left( \log P_{ii} + \log Q_{ii} \right)
\end{equation}

This loss pulls matched CAD--text pairs closer together while pushing unmatched pairs farther apart in the shared space. During this stage, we optimize the BrepEncoder to produce these language-aligned representations, while the entire CLIP text encoder remains frozen. This process yields a language-aligned geometric embedding that serves as the foundation for the subsequent LLM fine-tuning stages.


\subsection{Two-stage LLM Fine-tuning}
After aligning geometric and textual representations, we integrate the pretrained BrepEncoder into a large language model\cite{qwen3technicalreport} to form an end-to-end generation pipeline from geometry to text. The overall architecture comprises the B-rep encoding module, the geometry$\rightarrow$Q-Former projection, the pretrained Q-Former, and the LLM backbone. The BrepEncoder outputs a sequence of node tokens, denoted as $[h_1, h_2, \dots, h_n] \in \mathbb{R}^{N \times D_g}$. To match the pretrained BLIP-2 Q-Former embedding width ($D_{qf}=1408$), we use a two-layer MLP to project the 128-dimensional geometric tokens to $D_{qf}$, yielding the Q-Former value input $\mathbf{X}_{\text{qf}}\in\mathbb{R}^{T\times D_{qf}}$.

\noindent\textbf{Stage I: Geometry-to-Vision Bridging.}
We freeze the BrepEncoder, the Q-Former, and the LLM, and train only the geometry$\rightarrow$Q-Former projection head (two-layer MLP). To transform the variable-length geometric tokens into a fixed-size representation for stable training, the pretrained Q-Former employs $Q=32$ learnable query vectors. 
These queries cross-attend to the geometric embedding sequence $\mathbf{X}$, distilling the variable-length input into 32 fixed tokens. 
The resulting outputs are linearly mapped into the LLM hidden space and trained with an autoregressive cross-entropy objective. 
This establishes an initial semantic bridge from geometric tokens to the pretrained vision--language interface while preserving linguistic priors.

\noindent\textbf{Stage II: 3D--Language Alignment Fine-tuning.}
Building on Stage~I, we apply LoRA-based joint tuning to selected components: key sublayers and necessary normalization layers of the Q-Former, the geometry$\rightarrow$Q-Former projection, and a small subset of LLM parameters. The BrepEncoder remains frozen. 
This stage aims to establish a stable and semantically coherent mapping between structured B-rep representations and the language space. By jointly adapting the cross-modal interface and the language backbone with a small learning rate, the model progressively acquires the capability to interpret Brep-specific geometric and topological patterns and express them in natural language.

\section{Brep2Text Dataset}
To train and evaluate our model, we introduce \textbf{Brep2Text}, the first large-scale instruction-tuning dataset and benchmark specifically designed for the B-rep data modality. Built upon the Text2CAD corpus, Brep2Text leverages two semantic tiers: the abstract level and the beginner level. The abstract level prompts high-level semantic descriptions of CAD objects (e.g., function or category), while the beginner level focuses on outlining basic modeling steps. This dual-tier design explicitly encourages the model to jointly acquire categorical semantic understanding and procedural modeling logic—two core capabilities essential for real-world CAD interaction.

For each of the 134,722 unique B-rep models, we use Qwen-Max to automatically generate level-specific questions that are semantically aligned with the original human-written descriptions in Text2CAD, which serve as ground-truth answers. This yields a total of 269,444 high-quality question–answer pairs. Furthermore, we strictly reserve 200 CAD models as a held-out test set to ensure reliable evaluation. Brep2Text is not only the first instruction-following dataset tailored to B-rep representations but also establishes a reproducible benchmark for parametric CAD understanding tasks. See Appendix for details.

\section{Experiment}
\subsection{Experimental Setup}

\textbf{Implementation Details.} We adopt a two-step training scheme and conduct all experiments on 4 * 80GB A800 GPUs. We select the Qwen3-8B~\cite{qwen3technicalreport} model as the large language model backbone and initialize the Q-Former and projector using the pre-trained weights of TinyGPT-V~\cite{yuan2024tinygptv}. In the \textbf{first stage}, we utilize the AdamW optimizer with an initial learning rate of $1 \times 10^{-4}$ and a weight decay of 0.01. A linear warmup is applied for the first 5 epochs, followed by a cosine annealing strategy. The batch size is set to 256, and the model is trained for a total of 200 epochs with mixed-precision training and gradient clipping enabled. In the \textbf{second stage}, the weight decay is set to 0.05. The learning rate for the stage-1 is set to $3 \times 10^{-5}$, and $5 \times 10^{-6}$ for the stage-2. Following the settings of previous works~\cite{pointllm,shapellm,minigpt_3d}, we allocate 200 samples to the test set. For a fair comparison with point cloud-based baseline methods, each B-rep model is converted into a point cloud representation $P \in \mathbb{R}^{n \times d}$, where the number of points $n$ is 8192 and the feature dimension $d$ is 6 (missing color channels are set to black). For the VLM baselines, we render standard three-view images for each B-rep model. Additionally, we employ the ``Qwen3-30B-A3B-Instruct-2507'' model as the judge model during the evaluation phase. More implementation and training details can be found in the supplementary material.

\noindent\textbf{Baselines.} To investigate the performance gap among image-based, point cloud-based, and B-rep-based Multimodal Large Language Models, and to demonstrate the superiority of the B-rep data format in representing CAD geometry, we select several competitive 2D and 3D MLLMs as baselines. Under a consistent data split and evaluation protocol, we retrained all baselines on the Brep2Text dataset and uniformly use Qwen3-30B~\cite{qwen3technicalreport} for objective quantitative evaluation.


\subsection{3D Object Captioning}
To evaluate the model's fine-grained semantic understanding of 3D CAD objects, we conduct a 3D object captioning task on the Brep2Text dataset.

\textbf{Settings.} All models are prompted with a unified instruction: ``How was this CAD model constructed? Please describe this CAD model in detail.'' To ensure a comprehensive evaluation, we adopt three complementary assessment protocols: (1) Large Language Model evaluation, where Qwen3-30B judges the semantic consistency between the generated captions and human annotations; (2) Traditional metric evaluation, following the methods of PointLLM~\cite{pointllm} and MiniGPT-3D~\cite{minigpt_3d}, which uses Sentence-BERT~\cite{reimers2019sentencebertsentenceembeddingsusing} and SimCSE~\cite{gao2022simcsesimplecontrastivelearning} to compute embedding-based semantic similarity; (3) Human evaluation, where volunteers perform blind assessments using FreeCAD. Volunteers visually inspect each B-rep model and score the models' responses along two dimensions: \textit{Correctness} (the number of accurately described attributes, such as shape, dimensions, and modeling steps) and \textit{Hallucination} (the count of unsupported or fabricated details). \textit{Precision} is defined as the ratio of the number of correct attributes to the total number of asserted attributes.

\textbf{Results.} As shown in Table~\ref{tab:caption_exp}, BrepLLM achieves state-of-the-art performance across all evaluation dimensions. It obtains a Qwen3-30B score of 61.36, outperforming the strongest baseline (MiniGPT-3D) by 4.78 points. Furthermore, it improves the Sentence-BERT and SimCSE similarity scores by 3.78 and 4.10, respectively, demonstrating its superior understanding of fine-grained geometric and parametric structures. In human evaluation, BrepLLM achieves a precision of 83.62 and the lowest hallucination rate of 0.83, surpassing all baselines in generating accurate and structurally faithful descriptions. It is worth mentioning that our 8.4B BrepLLM comprehensively outperforms larger 13B-parameter models in both automatic and human evaluations. These results validate BrepLLM's strong capability to capture CAD-specific semantics and faithfully reflect them, showcasing the fine-grained 3D understanding inherited from B-rep representation learning.


\begin{table}[!t]
\centering
\caption{3D object captioning results on the Brep2Text dataset. Results are reported across human evaluation, LLM evaluation, and traditional metrics. The \textbf{bold} and \underline{underline} denote the best and second best, respectively; green numbers show gains over the second best.}
\label{tab:caption_exp}
\small
\setlength{\tabcolsep}{5pt}
\renewcommand{\arraystretch}{1.05} 
\resizebox{\textwidth}{!}{%
\begin{tabular}{l c c c c c c c}
\toprule
\multirow{2}{*}{Model} &
\multirow{2}{*}{LLM Size} &
\multirow{2}{*}{Qwen3-30B} &
\multirow{2}{*}{Sentence-BERT} &
\multirow{2}{*}{SimCSE} &
\multicolumn{3}{c}{Human Evaluation} \\
\cmidrule(lr){6-8}
& & & & & Correctness & Hallucination$\downarrow$ & Precision \\
\midrule
LLaVA-13B~\cite{liu2023visualinstructiontuning}       & 13B  & 36.73 & 45.67 & 47.16 & 2.16 & 1.58 & 63.15 \\
Qwen3-VL-8B~\cite{bai2025qwen3vltechnicalreport}     & 8B   & 55.48 & 67.65 & 68.85 & 3.97 & \underline{0.86} & 78.76 \\
\bottomrule
PointLLM-7B~\cite{xu2024pointllmempoweringlargelanguage}     & 7B   & 46.81 & 65.72 & 66.05 & 3.32 & 1.13 & 74.60 \\
PointLLM-13B~\cite{xu2024pointllmempoweringlargelanguage}    & 13B  & 49.65 & 66.78 & 67.32 & 3.46 & 1.28 & 72.99 \\
ShapeLLM-7B~\cite{qi2024shapellmuniversal3dobject}     & 7B   & 48.32 & 67.14 & 68.77 & 3.79 & 0.96 & \underline{79.78} \\
ShapeLLM-13B~\cite{qi2024shapellmuniversal3dobject}    & 13B  & 51.36 & 68.36 & 70.12 & 3.74 & 1.35 & 73.47 \\
Minigpt-3D~\cite{tang2024minigpt3defficientlyaligning3d}      & \textbf{2.7B} & \underline{56.58} & \underline{71.64} & \underline{73.13} & \underline{4.01} & 1.04 & 79.40 \\

\rowcolor{gray!10}
& &
\textbf{61.36} & \textbf{75.42} & \textbf{77.23} & \textbf{4.63} & \textbf{0.83} & \textbf{83.62} \\
\rowcolor{gray!10}
\multirow{-2}{*}{\textbf{BrepLLM}} & 
\multirow{-2}{*}{8.4B} &
\textcolor{green!50!black}{(+4.78)} & 
\textcolor{green!50!black}{(+3.78)} & 
\textcolor{green!50!black}{(+4.10)} & 
\textcolor{green!50!black}{(+0.62)} & 
\textcolor{green!50!black}{(+0.03)} & 
\textcolor{green!50!black}{(+3.84)} \\

\bottomrule
\end{tabular}%
}
\vspace{-0.6em}
\end{table}

\begin{table}[!t]
\centering
\caption{Generative 3D object classification on the Brep2Text test split. We report accuracy under two prompts (I/C) and their arithmetic mean.}
\label{tab:g3dcls_singlecol}

\scalebox{0.8}{%
\footnotesize
\setlength{\tabcolsep}{6pt}        
\renewcommand{\arraystretch}{1.05} 

\begin{tabular}{l c c c c c}
\toprule
Model & LLM Size & Input & I (\%) & C (\%) & Average \\
\midrule
LLaVA-13B~\cite{liu2023visualinstructiontuning}        & 13B  & Single-V. Img. & 46.5 & 44.5 & 45.5 \\
Qwen3-VL-8B~\cite{bai2025qwen3vltechnicalreport}      & 8B   & Tri-V. Img. & 54.0 & 55.0 & 54.5 \\
\midrule
PointLLM-7B~\cite{xu2024pointllmempoweringlargelanguage}     & 7B   & 3D Point Cloud & 52.5 & 51.0 & 51.75 \\
PointLLM-13B~\cite{xu2024pointllmempoweringlargelanguage}    & 13B  & 3D Point Cloud & 53.0 & 51.5 & 52.25 \\
ShapeLLM-7B~\cite{qi2024shapellmuniversal3dobject}     & 7B   & 3D Point Cloud & 53.5 & 52.5 & 53.0 \\
ShapeLLM-13B~\cite{qi2024shapellmuniversal3dobject}    & 13B  & 3D Point Cloud & 54.5 & 53.0 & 53.75 \\
Minigpt-3D~\cite{tang2024minigpt3defficientlyaligning3d}      & \textbf{2.7B} & 3D Point Cloud & \underline{56.0} & \underline{56.5} & \underline{56.25} \\
\rowcolor{gray!10}
\textbf{BrepLLM} & 8.4B & \textbf{B-rep} &
\makecell[c]{\textbf{60.5}\\[-0.2ex]\textcolor{green!50!black}{(+4.5)}} &
\makecell[c]{\textbf{59.5}\\[-0.2ex]\textcolor{green!50!black}{(+3.0)}} &
\makecell[c]{\textbf{60.0}\\[-0.2ex]\textcolor{green!50!black}{(+3.75)}} \\
\bottomrule
\end{tabular}%
}

\vspace{-0.3cm}
\end{table}

\subsection{Generative 3D Object Classification}
To evaluate BrepLLM's categorical understanding of parametric 3D CAD geometry, we formulate a generative object classification task based on free-form text generation on the Brep2Text test split.

\textbf{Settings.} Each sample is queried using two prompt styles: instructional (I) ``What is this?'', and completion (C) ``This is an object of ''. Given the B-rep input and a prompt, the model generates a text response. Following the evaluation protocol of the captioning task, we employ Qwen3-30B as an automatic judge to determine whether the generated response matches the ground-truth category. We report the classification accuracy (\%) under the instructional (I) and completion (C) settings, as well as their arithmetic mean (\emph{Average}).

\textbf{Results.} As shown in Table~\ref{tab:g3dcls_singlecol}, BrepLLM achieves an accuracy of 60.5\% under instructional prompting (I) and 59.5\% under completion prompting (C), resulting in an average accuracy of 60.0\%. This result comprehensively outperforms the baselines, exceeding them by margins of +4.5, +3.0, and +3.75 percentage points on the I, C, and \emph{Average} metrics, respectively. Notably, BrepLLM exhibits highly consistent performance across different prompt formats---the gap between I and C is only 1.0 percentage point, reflecting its robustness to prompt variations. Furthermore, our 8.4B-parameter BrepLLM surpasses all larger 13B-parameter baselines. This superiority stems from its direct modeling of B-rep structures, which explicitly encodes geometric and topological information, thereby enabling reliable category inference even in open-ended generative scenarios.


\subsection{Ablation Studies}
Under the same settings and evaluation protocol as the main experiments, we conduct ablation studies on the core components to quantify their necessity and contribution to downstream task performance.

\textbf{Adaptive UV sampling.} As shown in Table~\ref{tab:abl_uv_main_effect}, adaptive UV sampling brings a significant performance improvement of +3.24\% in the first stage (Stage~I). Although the relative gain slightly narrows after full training (Stage~I+II), it still maintains a stable contribution of +2.15\%. This suggests that its primary role is to enhance the model's early perception of fine-grained geometric structures. Even as the model's overall semantic understanding capabilities improve, adaptive UV sampling remains a crucial module for capturing precise surface details.

\textbf{Hierarchical BrepEncoder.} As shown in Table~\ref{tab:abl_hfe_main_effect}, introducing hierarchical feature extraction brings significant and consistent improvements across all training stages, increasing accuracy by +4.19\% in Stage~I and +3.97\% in Stage~I+II. This indicates that explicitly modeling the hierarchical structure of B-reps can effectively capture multi-granular geometric information, thereby significantly enhancing the model's deep semantic understanding of complex CAD objects.

\textbf{Training process.} As shown in Table~\ref{tab:abl_trainproc}, we validate the proposed two-stage fine-tuning scheme. When only Stage~I is enabled, the accuracy is 43.83\%, indicating that while the initial geometry-to-vision mapping is helpful, it is insufficient to support high-quality language generation. When training with only Stage~II, the accuracy reaches 59.12\%, demonstrating that 3D-language alignment is the primary driver of performance improvement. However, skipping Stage~I limits the model's geometric prior knowledge. Activating both stages simultaneously achieves the best performance with an accuracy of 61.36\%. This demonstrates that adopting a progressive training strategy can more fully unleash the model's potential.

\begin{table}[t]
  \centering
  \scriptsize
  \setlength{\tabcolsep}{3.0pt} 
  \renewcommand{\arraystretch}{1}

  \begin{minipage}[t]{0.32\textwidth} 
    \centering
    \caption{Adaptive UV sampling.}
    \label{tab:abl_uv_main_effect}
    \vspace{0.1cm}
    \begin{tabular}[t]{lccc} 
      \toprule
      Stage & w/o (\%) & w/ (\%) & $\Delta$ \\
      \midrule
      I            & 40.59 & 43.83 & +3.24 \\
      I+II         & 59.21 & 61.36 & +2.15 \\
      \vphantom{$\checkmark$} & & & \\ 
      \bottomrule
    \end{tabular}
  \end{minipage}\hfill
  \begin{minipage}[t]{0.33\textwidth} 
    \centering
    \caption{Hierarchical BrepEncoder.}
    \label{tab:abl_hfe_main_effect}
    \vspace{0.1cm}
    \begin{tabular}[t]{lccc} 
      \toprule
      Stage & w/o (\%) & w/ (\%) & $\Delta$ \\
      \midrule
      I            & 39.64 & 43.83 & +4.19 \\
      I+II         & 57.39 & 61.36 & +3.97 \\
      \vphantom{$\checkmark$} & & & \\ 
      \bottomrule
    \end{tabular}
  \end{minipage}\hfill
  \begin{minipage}[t]{0.28\textwidth} 
    \centering
    \caption{Training process.}
    \label{tab:abl_trainproc}
    \vspace{0.1cm}
    \begin{tabular}[t]{ccc} 
      \toprule
      I & II  & Accuracy (\%) \\
      \midrule
      $\checkmark$ &              & 43.83 \\
                   & $\checkmark$ & 59.12 \\
      $\checkmark$ & $\checkmark$ & 61.36 \\
      \bottomrule
    \end{tabular}
  \end{minipage}
\end{table}



\begin{table}[!t]
\centering
\caption{Qualitative comparison for 3D object captioning on Brep2Text dataset against the next-best baseline. Green highlights indicate salient details that align with the ground truth.}
\label{tab:qual_compare}
\footnotesize
\setlength{\tabcolsep}{4pt}
\renewcommand{\arraystretch}{1.08}

\begin{tabular}{>{\raggedright\arraybackslash}p{0.16\linewidth}
                >{\raggedright\arraybackslash}p{0.39\linewidth}
                >{\raggedright\arraybackslash}p{0.39\linewidth}}
\toprule
\textbf{Samples~1,~2}
&
\multicolumn{1}{c}{\includegraphics[height=1.25cm]{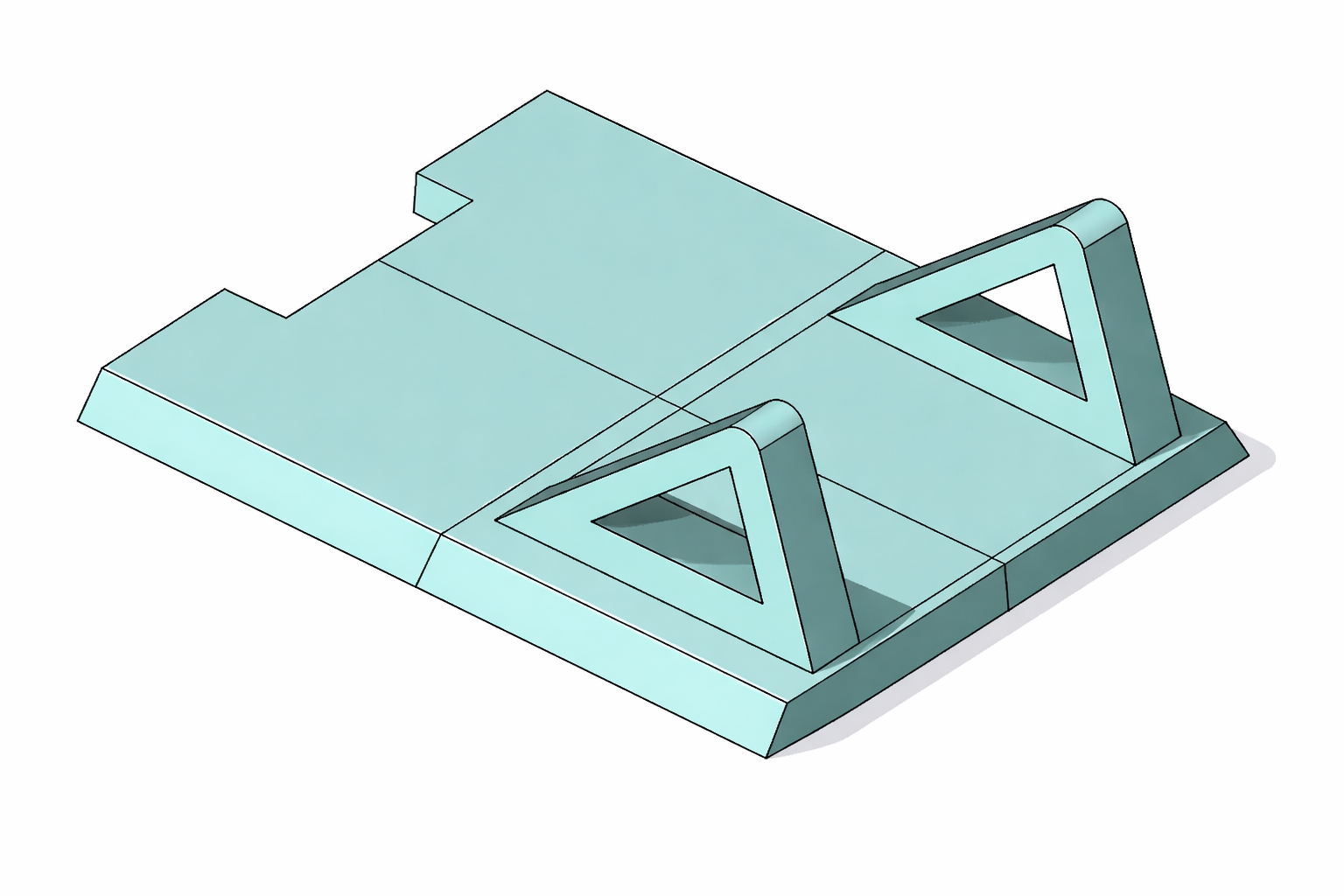}} &
\multicolumn{1}{c}{\includegraphics[height=1.25cm]{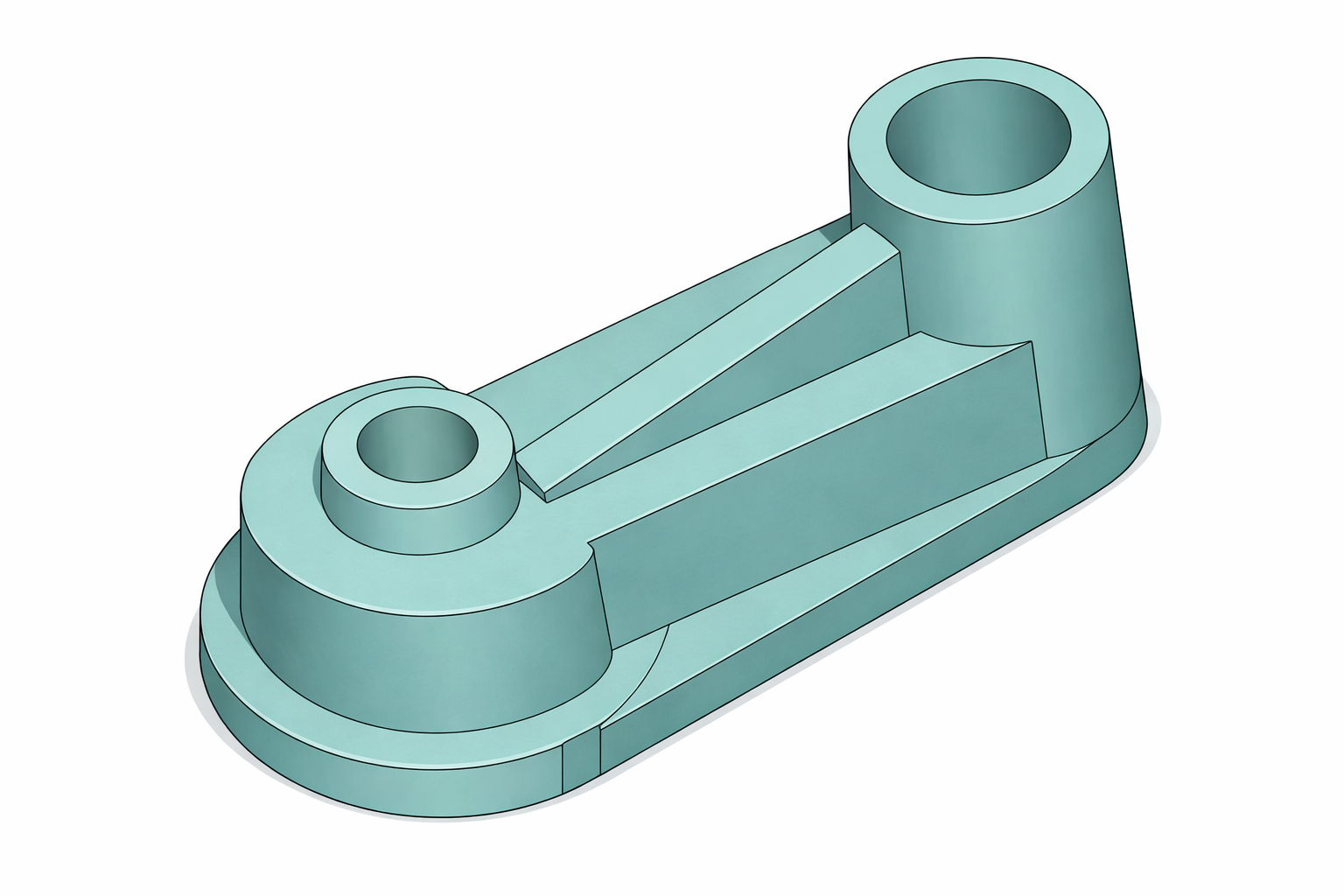}} \\
\midrule

\textbf{Prompt} &
\emph{What is this?} &
\emph{How this CAD model was constructed, describe this CAD model in detail.} \\

\textbf{Ground Truth} &
This is a rectangular plate with two triangular braces on its front face and a flat back; both side edges are beveled and slope upward, and there is a centered rectangular notch along the bottom edge. &
The design features a rounded rectangular base, a larger cylindrical body, a smaller cylindrical body, a cylindrical shape with a rectangular extension and a central hole, and a rectangular prism with a cylindrical base. \\
\midrule

MiniGPT-3D &
This is a rectangular CAD model with two triangular brackets on each side, integrated into the rectangular plate plane and with a cut at the bottom.
&
This model consists of a rectangular base, a large cylinder, and a small cylinder. It is made by stretching a rectangular base and adding circular features on both sides.
\\
\textbf{BrepLLM} &
\textbf{This is a CAD model based on a \textcolor{green!50!black}{flat rectangular plate}. It features a centered \textcolor{green!50!black}{rectangular notch} on the bottom edge and a flat back. The front face has two \textcolor{green!50!black}{symmetrically positioned} triangular braces \textcolor{green!50!black}{with internal triangular cutouts},and both side edges are slanted faces.}
&
\textbf{The model is built on a \textcolor{green!50!black}{rectangular base with rounded corners}. It contains \textcolor{green!50!black}{a larger cylinder} and \textcolor{green!50!black}{a smaller cylinder} at \textcolor{green!50!black}{each end}. It combines a \textcolor{green!50!black}{cylindrical body} with a rectangular extension and a \textcolor{green!50!black}{central hole}, as well as a straight prism with a \textcolor{green!50!black}{cylindrical base}.}\\
\bottomrule
\end{tabular}
\end{table}

\subsection{Qualitative Results}
Figure~\ref{fig:brepllm_interaction} qualitatively demonstrates BrepLLM's strong capability in perceiving and reasoning about CAD data. Through natural language interaction, BrepLLM can generate technically accurate descriptions of CAD models. It not only identifies the overall shape but also extracts specific dimensional information (e.g., length, width, and extrusion thickness), and clearly explains complex sequences of modeling operations, such as Extruded Cut and Circular Pattern.

Furthermore, Table~\ref{tab:qual_compare} presents a comparison between BrepLLM and the baseline on 3D object captioning. As shown in Sample~1, when faced with complex geometries, the baseline's description is relatively vague, whereas BrepLLM successfully identifies fine-grained part details, such as beveled side edges and internal triangular cutouts, providing a technically precise description that highly aligns with the ground truth. In Sample~2, which focuses on the model's construction method, BrepLLM accurately parses the topological structure of the model. It not only generates a comprehensive geometric description but also precisely localizes the spatial composition of various components, such as a rectangular base with rounded corners and large and small cylinders located at both ends. Its grasp of detail is significantly superior to that of the baseline. More qualitative results and comparisons with other models are provided in the Appendix.

\section{Conclusions and Future Work}
\label{sec:conclusion}

We introduced BrepLLM, the first multimodal framework that enables LLMs to directly parse and reason over raw B-rep data. The framework employs a two-stage training architecture comprising cross-modal alignment pre-training and LLM fine-tuning. It generates compact tokens fused with geometric and topological information and seamlessly integrates them into the LLM, thereby achieving an in-depth understanding of 3D geometry.

Our main contributions include: (i) an adaptive UV sampling strategy to construct graph representations that fuse geometric and topological information; (ii) a hierarchical BrepEncoder (face/edge/global topology) for extracting global and node tokens; (iii) a CLIP-style contrastive learning approach to align B-rep features into a unified geometry-language space; (iv) a two-stage progressive 3D-language semantic alignment strategy; and (v) Brep2Text, a large-scale benchmark dataset specifically designed for B-rep understanding.

This study demonstrates that processing native B-reps directly enhances LLMs' geometric and topological understanding. Future work will scale up high-quality multimodal B-rep data to improve generalization. We also plan to extend this framework to B-rep generation, editing, and assembly understanding, exploring its potential in industrial intelligent CAD workflows.



%
%
\bibliographystyle{splncs04}
\bibliography{main}

@String(CVPR  = {IEEE Conf. Comput. Vis. Pattern Recog.})

@String(NeurIPS = {Adv. Neural Inform. Process. Syst.})

@String(AAAI  = {AAAI})

@String(CVPR  = {CVPR})

@String(NeurIPS = {NeurIPS})

@article{query2cad,
  title={Query2cad: Generating cad models using natural language queries},
  author={Badagabettu, Akshay and Yarlagadda, Sai Sravan and Farimani, Amir Barati},
  journal={arXiv preprint arXiv:2406.00144},
  year={2024}
}

@article{text2cad,
  title={Text2cad: Generating sequential cad designs from beginner-to-expert level text prompts},
  author={Khan, Mohammad Sadil and Sinha, Sankalp and Uddin, Talha and Stricker, Didier and Ali, Sk Aziz and Afzal, Muhammad Zeshan},
  journal={Advances in Neural Information Processing Systems},
  volume={37},
  pages={7552--7579},
  year={2024}
}

@article{CAD-Editor,
  title={CAD-Editor: Text-based CAD Editing through Adapting Large Language Models with Synthetic Data},
  author={Yuan, Yu and Sun, Shizhao and Liu, Qi and Bian, Jiang}
}

@article{Cad-assistant,
  title={Cad-assistant: Tool-augmented vllms as generic cad task solvers},
  author={Mallis, Dimitrios and Karadeniz, Ahmet Serdar and Cavada, Sebastian and Rukhovich, Danila and Foteinopoulou, Niki and Cherenkova, Kseniya and Kacem, Anis and Aouada, Djamila},
  journal={arXiv preprint arXiv:2412.13810},
  year={2024}
}

@inproceedings{cadvlm,
  title={Cadvlm: Bridging language and vision in the generation of parametric cad sketches},
  author={Wu, Sifan and Khasahmadi, Amir Hosein and Katz, Mor and Jayaraman, Pradeep Kumar and Pu, Yewen and Willis, Karl and Liu, Bang},
  booktitle={European Conference on Computer Vision},
  pages={368--384},
  year={2024},
  organization={Springer}
}

@inproceedings{llm4cad,
  title={LLM4CAD: Multi-Modal Large Language Models for 3D Computer-Aided Design Generation},
  author={Li, Xingang and Sun, Yuewan and Sha, Zhenghui},
  booktitle={International Design Engineering Technical Conferences and Computers and Information in Engineering Conference},
  volume={88407},
  pages={V006T06A015},
  year={2024},
  organization={American Society of Mechanical Engineers}
}

@article{Cad-mllm,
  title={Cad-mllm: Unifying multimodality-conditioned cad generation with mllm},
  author={Xu, Jingwei and Zhao, Zibo and Wang, Chenyu and Liu, Wen and Ma, Yi and Gao, Shenghua},
  journal={arXiv preprint arXiv:2411.04954},
  year={2024}
}

@inproceedings{Cad-gpt,
  title={Cad-gpt: Synthesising cad construction sequence with spatial reasoning-enhanced multimodal llms},
  author={Wang, Siyu and Chen, Cailian and Le, Xinyi and Xu, Qimin and Xu, Lei and Zhang, Yanzhou and Yang, Jie},
  booktitle={Proceedings of the AAAI Conference on Artificial Intelligence},
  volume={39},
  number={8},
  pages={7880--7888},
  year={2025}
}

@inproceedings{Cad-llm,
  title={Cad-llm: Large language model for cad generation},
  author={Wu, Sifan and Khasahmadi, Amir and Katz, Mor and Jayaraman, Pradeep Kumar and Pu, Yewen and Willis, Karl and Liu, Bang},
  booktitle={Proceedings of the neural information processing systems conference. neurIPS},
  volume={1},
  year={2023}
}

@misc{minigpt_3d,
      title={MiniGPT-3D: Efficiently Aligning 3D Point Clouds with Large Language Models using 2D Priors}, 
      author={Yuan Tang and Xu Han and Xianzhi Li and Qiao Yu and Yixue Hao and Long Hu and Min Chen},
      year={2024},
      eprint={2405.01413},
      archivePrefix={arXiv},
      primaryClass={cs.CV},
      url={https://arxiv.org/abs/2405.01413}, 
}

@inproceedings{clip,
  title={Learning transferable visual models from natural language supervision},
  author={Radford, Alec and Kim, Jong Wook and Hallacy, Chris and Ramesh, Aditya and Goh, Gabriel and Agarwal, Sandhini and Sastry, Girish and Askell, Amanda and Mishkin, Pamela and Clark, Jack and others},
  booktitle={International conference on machine learning},
  pages={8748--8763},
  year={2021},
  organization={PmLR}
}

@misc{pointllm,
      title={PointLLM: Empowering Large Language Models to Understand Point Clouds}, 
      author={Runsen Xu and Xiaolong Wang and Tai Wang and Yilun Chen and Jiangmiao Pang and Dahua Lin},
      year={2024},
      eprint={2308.16911},
      archivePrefix={arXiv},
      primaryClass={cs.CV},
      url={https://arxiv.org/abs/2308.16911}, 
}

@misc{shapellm,
      title={ShapeLLM: Universal 3D Object Understanding for Embodied Interaction}, 
      author={Zekun Qi and Runpei Dong and Shaochen Zhang and Haoran Geng and Chunrui Han and Zheng Ge and Li Yi and Kaisheng Ma},
      year={2024},
      eprint={2402.17766},
      archivePrefix={arXiv},
      primaryClass={cs.CV},
      url={https://arxiv.org/abs/2402.17766}, 
}

@article{Wu2023PointTV,
  title={Point Transformer V3: Simpler, Faster, Stronger},
  author={Xiaoyang Wu and Li Jiang and Peng-Shuai Wang and Zhijian Liu and Xihui Liu and Yu Qiao and Wanli Ouyang and Tong He and Hengshuang Zhao},
  journal={2024 IEEE/CVF Conference on Computer Vision and Pattern Recognition (CVPR)},
  year={2023},
  pages={4840-4851},
  url={https://api.semanticscholar.org/CorpusID:266335894}
}

@misc{kolodiazhnyi2026cadrillemultimodalcadreconstruction,
      title={cadrille: Multi-modal CAD Reconstruction with Reinforcement Learning}, 
      author={Maksim Kolodiazhnyi and Denis Tarasov and Dmitrii Zhemchuzhnikov and Alexander Nikulin and Ilya Zisman and Anna Vorontsova and Anton Konushin and Vladislav Kurenkov and Danila Rukhovich},
      year={2026},
      eprint={2505.22914},
      archivePrefix={arXiv},
      primaryClass={cs.CV},
      url={https://arxiv.org/abs/2505.22914}, 
}

@misc{xu2024brepgenbrepgenerativediffusion,
      title={BrepGen: A B-rep Generative Diffusion Model with Structured Latent Geometry}, 
      author={Xiang Xu and Joseph G. Lambourne and Pradeep Kumar Jayaraman and Zhengqing Wang and Karl D. D. Willis and Yasutaka Furukawa},
      year={2024},
      eprint={2401.15563},
      archivePrefix={arXiv},
      primaryClass={cs.CV},
      url={https://arxiv.org/abs/2401.15563}, 
}

@misc{openai2024gpt4technicalreport,
      title={GPT-4 Technical Report}, 
      author={OpenAI and Josh Achiam and Steven Adler and Sandhini Agarwal and Lama Ahmad and Ilge Akkaya and Florencia Leoni Aleman and Diogo Almeida and Janko Altenschmidt and Sam Altman and Shyamal Anadkat and Red Avila and Igor Babuschkin and Suchir Balaji and Valerie Balcom and Paul Baltescu and Haiming Bao and Mohammad Bavarian and Jeff Belgum and Irwan Bello and Jake Berdine and Gabriel Bernadett-Shapiro and Christopher Berner and Lenny Bogdonoff and Oleg Boiko and Madelaine Boyd and Anna-Luisa Brakman and Greg Brockman and Tim Brooks and Miles Brundage and Kevin Button and Trevor Cai and Rosie Campbell and Andrew Cann and Brittany Carey and Chelsea Carlson and Rory Carmichael and Brooke Chan and Che Chang and Fotis Chantzis and Derek Chen and Sully Chen and Ruby Chen and Jason Chen and Mark Chen and Ben Chess and Chester Cho and Casey Chu and Hyung Won Chung and Dave Cummings and Jeremiah Currier and Yunxing Dai and Cory Decareaux and Thomas Degry and Noah Deutsch and Damien Deville and Arka Dhar and David Dohan and Steve Dowling and Sheila Dunning and Adrien Ecoffet and Atty Eleti and Tyna Eloundou and David Farhi and Liam Fedus and Niko Felix and Simón Posada Fishman and Juston Forte and Isabella Fulford and Leo Gao and Elie Georges and Christian Gibson and Vik Goel and Tarun Gogineni and Gabriel Goh and Rapha Gontijo-Lopes and Jonathan Gordon and Morgan Grafstein and Scott Gray and Ryan Greene and Joshua Gross and Shixiang Shane Gu and Yufei Guo and Chris Hallacy and Jesse Han and Jeff Harris and Yuchen He and Mike Heaton and Johannes Heidecke and Chris Hesse and Alan Hickey and Wade Hickey and Peter Hoeschele and Brandon Houghton and Kenny Hsu and Shengli Hu and Xin Hu and Joost Huizinga and Shantanu Jain and Shawn Jain and Joanne Jang and Angela Jiang and Roger Jiang and Haozhun Jin and Denny Jin and Shino Jomoto and Billie Jonn and Heewoo Jun and Tomer Kaftan and Łukasz Kaiser and Ali Kamali and Ingmar Kanitscheider and Nitish Shirish Keskar and Tabarak Khan and Logan Kilpatrick and Jong Wook Kim and Christina Kim and Yongjik Kim and Jan Hendrik Kirchner and Jamie Kiros and Matt Knight and Daniel Kokotajlo and Łukasz Kondraciuk and Andrew Kondrich and Aris Konstantinidis and Kyle Kosic and Gretchen Krueger and Vishal Kuo and Michael Lampe and Ikai Lan and Teddy Lee and Jan Leike and Jade Leung and Daniel Levy and Chak Ming Li and Rachel Lim and Molly Lin and Stephanie Lin and Mateusz Litwin and Theresa Lopez and Ryan Lowe and Patricia Lue and Anna Makanju and Kim Malfacini and Sam Manning and Todor Markov and Yaniv Markovski and Bianca Martin and Katie Mayer and Andrew Mayne and Bob McGrew and Scott Mayer McKinney and Christine McLeavey and Paul McMillan and Jake McNeil and David Medina and Aalok Mehta and Jacob Menick and Luke Metz and Andrey Mishchenko and Pamela Mishkin and Vinnie Monaco and Evan Morikawa and Daniel Mossing and Tong Mu and Mira Murati and Oleg Murk and David Mély and Ashvin Nair and Reiichiro Nakano and Rajeev Nayak and Arvind Neelakantan and Richard Ngo and Hyeonwoo Noh and Long Ouyang and Cullen O'Keefe and Jakub Pachocki and Alex Paino and Joe Palermo and Ashley Pantuliano and Giambattista Parascandolo and Joel Parish and Emy Parparita and Alex Passos and Mikhail Pavlov and Andrew Peng and Adam Perelman and Filipe de Avila Belbute Peres and Michael Petrov and Henrique Ponde de Oliveira Pinto and Michael and Pokorny and Michelle Pokrass and Vitchyr H. Pong and Tolly Powell and Alethea Power and Boris Power and Elizabeth Proehl and Raul Puri and Alec Radford and Jack Rae and Aditya Ramesh and Cameron Raymond and Francis Real and Kendra Rimbach and Carl Ross and Bob Rotsted and Henri Roussez and Nick Ryder and Mario Saltarelli and Ted Sanders and Shibani Santurkar and Girish Sastry and Heather Schmidt and David Schnurr and John Schulman and Daniel Selsam and Kyla Sheppard and Toki Sherbakov and Jessica Shieh and Sarah Shoker and Pranav Shyam and Szymon Sidor and Eric Sigler and Maddie Simens and Jordan Sitkin and Katarina Slama and Ian Sohl and Benjamin Sokolowsky and Yang Song and Natalie Staudacher and Felipe Petroski Such and Natalie Summers and Ilya Sutskever and Jie Tang and Nikolas Tezak and Madeleine B. Thompson and Phil Tillet and Amin Tootoonchian and Elizabeth Tseng and Preston Tuggle and Nick Turley and Jerry Tworek and Juan Felipe Cerón Uribe and Andrea Vallone and Arun Vijayvergiya and Chelsea Voss and Carroll Wainwright and Justin Jay Wang and Alvin Wang and Ben Wang and Jonathan Ward and Jason Wei and CJ Weinmann and Akila Welihinda and Peter Welinder and Jiayi Weng and Lilian Weng and Matt Wiethoff and Dave Willner and Clemens Winter and Samuel Wolrich and Hannah Wong and Lauren Workman and Sherwin Wu and Jeff Wu and Michael Wu and Kai Xiao and Tao Xu and Sarah Yoo and Kevin Yu and Qiming Yuan and Wojciech Zaremba and Rowan Zellers and Chong Zhang and Marvin Zhang and Shengjia Zhao and Tianhao Zheng and Juntang Zhuang and William Zhuk and Barret Zoph},
      year={2024},
      eprint={2303.08774},
      archivePrefix={arXiv},
      primaryClass={cs.CL},
      url={https://arxiv.org/abs/2303.08774}, 
}

@misc{anil2023palm2technicalreport,
      title={PaLM 2 Technical Report}, 
      author={Rohan Anil and Andrew M. Dai and Orhan Firat and Melvin Johnson and Dmitry Lepikhin and Alexandre Passos and Siamak Shakeri and Emanuel Taropa and Paige Bailey and Zhifeng Chen and Eric Chu and Jonathan H. Clark and Laurent El Shafey and Yanping Huang and Kathy Meier-Hellstern and Gaurav Mishra and Erica Moreira and Mark Omernick and Kevin Robinson and Sebastian Ruder and Yi Tay and Kefan Xiao and Yuanzhong Xu and Yujing Zhang and Gustavo Hernandez Abrego and Junwhan Ahn and Jacob Austin and Paul Barham and Jan Botha and James Bradbury and Siddhartha Brahma and Kevin Brooks and Michele Catasta and Yong Cheng and Colin Cherry and Christopher A. Choquette-Choo and Aakanksha Chowdhery and Clément Crepy and Shachi Dave and Mostafa Dehghani and Sunipa Dev and Jacob Devlin and Mark Díaz and Nan Du and Ethan Dyer and Vlad Feinberg and Fangxiaoyu Feng and Vlad Fienber and Markus Freitag and Xavier Garcia and Sebastian Gehrmann and Lucas Gonzalez and Guy Gur-Ari and Steven Hand and Hadi Hashemi and Le Hou and Joshua Howland and Andrea Hu and Jeffrey Hui and Jeremy Hurwitz and Michael Isard and Abe Ittycheriah and Matthew Jagielski and Wenhao Jia and Kathleen Kenealy and Maxim Krikun and Sneha Kudugunta and Chang Lan and Katherine Lee and Benjamin Lee and Eric Li and Music Li and Wei Li and YaGuang Li and Jian Li and Hyeontaek Lim and Hanzhao Lin and Zhongtao Liu and Frederick Liu and Marcello Maggioni and Aroma Mahendru and Joshua Maynez and Vedant Misra and Maysam Moussalem and Zachary Nado and John Nham and Eric Ni and Andrew Nystrom and Alicia Parrish and Marie Pellat and Martin Polacek and Alex Polozov and Reiner Pope and Siyuan Qiao and Emily Reif and Bryan Richter and Parker Riley and Alex Castro Ros and Aurko Roy and Brennan Saeta and Rajkumar Samuel and Renee Shelby and Ambrose Slone and Daniel Smilkov and David R. So and Daniel Sohn and Simon Tokumine and Dasha Valter and Vijay Vasudevan and Kiran Vodrahalli and Xuezhi Wang and Pidong Wang and Zirui Wang and Tao Wang and John Wieting and Yuhuai Wu and Kelvin Xu and Yunhan Xu and Linting Xue and Pengcheng Yin and Jiahui Yu and Qiao Zhang and Steven Zheng and Ce Zheng and Weikang Zhou and Denny Zhou and Slav Petrov and Yonghui Wu},
      year={2023},
      eprint={2305.10403},
      archivePrefix={arXiv},
      primaryClass={cs.CL},
      url={https://arxiv.org/abs/2305.10403}, 
}

@misc{jiang2024mixtralexperts,
      title={Mixtral of Experts}, 
      author={Albert Q. Jiang and Alexandre Sablayrolles and Antoine Roux and Arthur Mensch and Blanche Savary and Chris Bamford and Devendra Singh Chaplot and Diego de las Casas and Emma Bou Hanna and Florian Bressand and Gianna Lengyel and Guillaume Bour and Guillaume Lample and Lélio Renard Lavaud and Lucile Saulnier and Marie-Anne Lachaux and Pierre Stock and Sandeep Subramanian and Sophia Yang and Szymon Antoniak and Teven Le Scao and Théophile Gervet and Thibaut Lavril and Thomas Wang and Timothée Lacroix and William El Sayed},
      year={2024},
      eprint={2401.04088},
      archivePrefix={arXiv},
      primaryClass={cs.LG},
      url={https://arxiv.org/abs/2401.04088}, 
}

@misc{li2023blip2bootstrappinglanguageimagepretraining,
      title={BLIP-2: Bootstrapping Language-Image Pre-training with Frozen Image Encoders and Large Language Models}, 
      author={Junnan Li and Dongxu Li and Silvio Savarese and Steven Hoi},
      year={2023},
      eprint={2301.12597},
      archivePrefix={arXiv},
      primaryClass={cs.CV},
      url={https://arxiv.org/abs/2301.12597}, 
}

@misc{bai2023qwenvlversatilevisionlanguagemodel,
      title={Qwen-VL: A Versatile Vision-Language Model for Understanding, Localization, Text Reading, and Beyond}, 
      author={Jinze Bai and Shuai Bai and Shusheng Yang and Shijie Wang and Sinan Tan and Peng Wang and Junyang Lin and Chang Zhou and Jingren Zhou},
      year={2023},
      eprint={2308.12966},
      archivePrefix={arXiv},
      primaryClass={cs.CV},
      url={https://arxiv.org/abs/2308.12966}, 
}

@misc{hong20233dllminjecting3dworld,
      title={3D-LLM: Injecting the 3D World into Large Language Models}, 
      author={Yining Hong and Haoyu Zhen and Peihao Chen and Shuhong Zheng and Yilun Du and Zhenfang Chen and Chuang Gan},
      year={2023},
      eprint={2307.12981},
      archivePrefix={arXiv},
      primaryClass={cs.CV},
      url={https://arxiv.org/abs/2307.12981}, 
}

@misc{xu2024pointllmempoweringlargelanguage,
      title={PointLLM: Empowering Large Language Models to Understand Point Clouds}, 
      author={Runsen Xu and Xiaolong Wang and Tai Wang and Yilun Chen and Jiangmiao Pang and Dahua Lin},
      year={2024},
      eprint={2308.16911},
      archivePrefix={arXiv},
      primaryClass={cs.CV},
      url={https://arxiv.org/abs/2308.16911}, 
}

@misc{guo2023pointbindpointllmaligning,
      title={Point-Bind \& Point-LLM: Aligning Point Cloud with Multi-modality for 3D Understanding, Generation, and Instruction Following}, 
      author={Ziyu Guo and Renrui Zhang and Xiangyang Zhu and Yiwen Tang and Xianzheng Ma and Jiaming Han and Kexin Chen and Peng Gao and Xianzhi Li and Hongsheng Li and Pheng-Ann Heng},
      year={2023},
      eprint={2309.00615},
      archivePrefix={arXiv},
      primaryClass={cs.CV},
      url={https://arxiv.org/abs/2309.00615}, 
}

@article{zhu2024llava,
  title={LLaVA-3D: A Simple yet Effective Pathway to Empowering LMMs with 3D-awareness},
  author={Zhu, Chenming and Wang, Tai and Zhang, Wenwei and Pang, Jiangmiao and Liu, Xihui},
  journal={arXiv preprint arXiv:2409.18125},
  year={2024}
}

@misc{tang2024minigpt3defficientlyaligning3d,
      title={MiniGPT-3D: Efficiently Aligning 3D Point Clouds with Large Language Models using 2D Priors}, 
      author={Yuan Tang and Xu Han and Xianzhi Li and Qiao Yu and Yixue Hao and Long Hu and Min Chen},
      year={2024},
      eprint={2405.01413},
      archivePrefix={arXiv},
      primaryClass={cs.CV},
      url={https://arxiv.org/abs/2405.01413}, 
}

@misc{heidari2025geometricdeeplearningcomputeraided,
      title={Geometric Deep Learning for Computer-Aided Design: A Survey}, 
      author={Negar Heidari and Alexandros Iosifidis},
      year={2025},
      eprint={2402.17695},
      archivePrefix={arXiv},
      primaryClass={cs.CG},
      url={https://arxiv.org/abs/2402.17695}, 
}

@misc{doris2025cadcoderopensourcevisionlanguagemodel,
      title={CAD-Coder: An Open-Source Vision-Language Model for Computer-Aided Design Code Generation}, 
      author={Anna C. Doris and Md Ferdous Alam and Amin Heyrani Nobari and Faez Ahmed},
      year={2025},
      eprint={2505.14646},
      archivePrefix={arXiv},
      primaryClass={cs.CV},
      url={https://arxiv.org/abs/2505.14646}, 
}

@misc{radford2021learningtransferablevisualmodels,
      title={Learning Transferable Visual Models From Natural Language Supervision}, 
      author={Alec Radford and Jong Wook Kim and Chris Hallacy and Aditya Ramesh and Gabriel Goh and Sandhini Agarwal and Girish Sastry and Amanda Askell and Pamela Mishkin and Jack Clark and Gretchen Krueger and Ilya Sutskever},
      year={2021},
      eprint={2103.00020},
      archivePrefix={arXiv},
      primaryClass={cs.CV},
      url={https://arxiv.org/abs/2103.00020}, 
}

@misc{qwen3technicalreport,
      title={Qwen3 Technical Report}, 
      author={Qwen Team},
      year={2025},
      eprint={2505.09388},
      archivePrefix={arXiv},
      primaryClass={cs.CL},
      url={https://arxiv.org/abs/2505.09388}, 
}

@misc{cadllama,
      title={CAD-Llama: Leveraging Large Language Models for Computer-Aided Design Parametric 3D Model Generation}, 
      author={Jiahao Li and Weijian Ma and Xueyang Li and Yunzhong Lou and Guichun Zhou and Xiangdong Zhou},
      year={2025},
      eprint={2505.04481},
      archivePrefix={arXiv},
      primaryClass={cs.CV},
      url={https://arxiv.org/abs/2505.04481}, 
}

@misc{flexcad,
      title={FlexCAD: Unified and Versatile Controllable CAD Generation with Fine-tuned Large Language Models}, 
      author={Zhanwei Zhang and Shizhao Sun and Wenxiao Wang and Deng Cai and Jiang Bian},
      year={2025},
      eprint={2411.05823},
      archivePrefix={arXiv},
      primaryClass={cs.CV},
      url={https://arxiv.org/abs/2411.05823}, 
}

@inproceedings{cadtranslator,
  title={{CAD Translator}: An Effective Drive for Text to 3D Parametric Computer-Aided Design Generative Modeling},
  author={Li, Jingen and others},
  booktitle={Proceedings of the 32nd ACM International Conference on Multimedia},
  year={2024},
  doi={10.1145/3664647.3681549},
  url={https://dl.acm.org/doi/10.1145/3664647.3681549}
}

@misc{blenderllm,
      title={BlenderLLM: Training Large Language Models for Computer-Aided Design with Self-improvement}, 
      author={Yuhao Du and Shunian Chen and Wenbo Zan and Peizhao Li and Mingxuan Wang and Dingjie Song and Bo Li and Yan Hu and Benyou Wang},
      year={2024},
      eprint={2412.14203},
      archivePrefix={arXiv},
      primaryClass={cs.HC},
      url={https://arxiv.org/abs/2412.14203}, 
}

@misc{greenplm,
      title={GreenPLM: Cross-Lingual Transfer of Monolingual Pre-Trained Language Models at Almost No Cost}, 
      author={Qingcheng Zeng and Lucas Garay and Peilin Zhou and Dading Chong and Yining Hua and Jiageng Wu and Yikang Pan and Han Zhou and Rob Voigt and Jie Yang},
      year={2023},
      eprint={2211.06993},
      archivePrefix={arXiv},
      primaryClass={cs.CL},
      url={https://arxiv.org/abs/2211.06993}, 
}

@misc{solidgen,
      title={SolidGen: An Autoregressive Model for Direct B-rep Synthesis}, 
      author={Pradeep Kumar Jayaraman and Joseph G. Lambourne and Nishkrit Desai and Karl D. D. Willis and Aditya Sanghi and Nigel J. W. Morris},
      year={2023},
      eprint={2203.13944},
      archivePrefix={arXiv},
      primaryClass={cs.LG},
      url={https://arxiv.org/abs/2203.13944}, 
}

@misc{neuronurbs,
      title={NeuroNURBS: Learning Efficient Surface Representations for 3D Solids}, 
      author={Jiajie Fan and Babak Gholami and Thomas Bäck and Hao Wang},
      year={2024},
      eprint={2411.10848},
      archivePrefix={arXiv},
      primaryClass={cs.CV},
      url={https://arxiv.org/abs/2411.10848}, 
}

@misc{dtgbrepgen,
      title={DTGBrepGen: A Novel B-rep Generative Model through Decoupling Topology and Geometry}, 
      author={Jing Li and Yihang Fu and Falai Chen},
      year={2025},
      eprint={2503.13110},
      archivePrefix={arXiv},
      primaryClass={cs.CV},
      url={https://arxiv.org/abs/2503.13110}, 
}

@article{hola,
  title={{HoLa}: B-Rep Generation using a Holistic Latent Representation},
  author={Liu, Yilin and Xu, Duoteng and Yu, Xingyao and Xu, Xiang and Cohen-Or, Daniel and Zhang, Hao and Huang, Hui},
  journal={ACM Transactions on Graphics},
  volume={44},
  number={4},
  pages={116},
  year={2025},
  doi={10.1145/3730842},
  eprint={2504.14257},
  archivePrefix={arXiv},
  url={https://arxiv.org/abs/2504.14257}
}

@inproceedings{autobrep,
  title={{AutoBrep}: Autoregressive B-Rep Generation with Unified Topology and Geometry},
  author={Xu, Xiang and Jayaraman, Pradeep Kumar and Lambourne, Joseph G. and Liu, Yilin and Malpure, Durvesh and Meltzer, Pete},
  booktitle={Proceedings of the SIGGRAPH Asia 2025 Conference Papers},
  pages={1--12},
  year={2025},
  doi={10.1145/3757377.3763814},
  eprint={2512.03018},
  archivePrefix={arXiv},
  url={https://arxiv.org/abs/2512.03018}
}

@inproceedings{abc,
  author={Koch, Sebastian and Matveev, Albert and Jiang, Zhongshi and Williams, Francis and Artemov, Alexey and Burnaev, Evgeny and Alexa, Marc and Zorin, Denis and Panozzo, Daniele},
  title={{ABC}: A Big {CAD} Model Dataset for Geometric Deep Learning},
  booktitle={Proceedings of the IEEE/CVF Conference on Computer Vision and Pattern Recognition (CVPR)},
  pages={9601--9611},
  year={2019},
  eprint={1812.06216},
  archivePrefix={arXiv},
  url={https://arxiv.org/abs/1812.06216}
}

@misc{xie2025texttocadquerynewparadigmcad,
      title={Text-to-CadQuery: A New Paradigm for CAD Generation with Scalable Large Model Capabilities}, 
      author={Haoyang Xie and Feng Ju},
      year={2025},
      eprint={2505.06507},
      archivePrefix={arXiv},
      primaryClass={cs.AI},
      url={https://arxiv.org/abs/2505.06507}, 
}

@misc{yuan2024tinygptv,
      title={TinyGPT-V: Efficient Multimodal Large Language Model via Small Backbones}, 
      author={Zhengqing Yuan and Zhaoxu Li and Weiran Huang and Yanfang Ye and Lichao Sun},
      year={2024},
      eprint={2312.16862},
      archivePrefix={arXiv},
      primaryClass={cs.CV}
}

@misc{reimers2019sentencebertsentenceembeddingsusing,
      title={Sentence-BERT: Sentence Embeddings using Siamese BERT-Networks}, 
      author={Nils Reimers and Iryna Gurevych},
      year={2019},
      eprint={1908.10084},
      archivePrefix={arXiv},
      primaryClass={cs.CL},
      url={https://arxiv.org/abs/1908.10084}, 
}

@misc{gao2022simcsesimplecontrastivelearning,
      title={SimCSE: Simple Contrastive Learning of Sentence Embeddings}, 
      author={Tianyu Gao and Xingcheng Yao and Danqi Chen},
      year={2022},
      eprint={2104.08821},
      archivePrefix={arXiv},
      primaryClass={cs.CL},
      url={https://arxiv.org/abs/2104.08821}, 
}

@misc{liu2023visualinstructiontuning,
      title={Visual Instruction Tuning}, 
      author={Haotian Liu and Chunyuan Li and Qingyang Wu and Yong Jae Lee},
      year={2023},
      eprint={2304.08485},
      archivePrefix={arXiv},
      primaryClass={cs.CV},
      url={https://arxiv.org/abs/2304.08485}, 
}

@misc{bai2025qwen3vltechnicalreport,
      title={Qwen3-VL Technical Report}, 
      author={Shuai Bai and Yuxuan Cai and Ruizhe Chen and Keqin Chen and Xionghui Chen and Zesen Cheng and Lianghao Deng and Wei Ding and Chang Gao and Chunjiang Ge and Wenbin Ge and Zhifang Guo and Qidong Huang and Jie Huang and Fei Huang and Binyuan Hui and Shutong Jiang and Zhaohai Li and Mingsheng Li and Mei Li and Kaixin Li and Zicheng Lin and Junyang Lin and Xuejing Liu and Jiawei Liu and Chenglong Liu and Yang Liu and Dayiheng Liu and Shixuan Liu and Dunjie Lu and Ruilin Luo and Chenxu Lv and Rui Men and Lingchen Meng and Xuancheng Ren and Xingzhang Ren and Sibo Song and Yuchong Sun and Jun Tang and Jianhong Tu and Jianqiang Wan and Peng Wang and Pengfei Wang and Qiuyue Wang and Yuxuan Wang and Tianbao Xie and Yiheng Xu and Haiyang Xu and Jin Xu and Zhibo Yang and Mingkun Yang and Jianxin Yang and An Yang and Bowen Yu and Fei Zhang and Hang Zhang and Xi Zhang and Bo Zheng and Humen Zhong and Jingren Zhou and Fan Zhou and Jing Zhou and Yuanzhi Zhu and Ke Zhu},
      year={2025},
      eprint={2511.21631},
      archivePrefix={arXiv},
      primaryClass={cs.CV},
      url={https://arxiv.org/abs/2511.21631}, 
}

@misc{qi2024shapellmuniversal3dobject,
      title={ShapeLLM: Universal 3D Object Understanding for Embodied Interaction}, 
      author={Zekun Qi and Runpei Dong and Shaochen Zhang and Haoran Geng and Chunrui Han and Zheng Ge and Li Yi and Kaisheng Ma},
      year={2024},
      eprint={2402.17766},
      archivePrefix={arXiv},
      primaryClass={cs.CV},
      url={https://arxiv.org/abs/2402.17766}, 
}

@misc{li2026autoregressivegenerationbrepholistic,
  title={AutoRegressive Generation with B-rep Holistic Token Sequence Representation},
  author={Li, Jiahao and Bai, Yunpeng and Dai, Yongkang and Guo, Hao and Gan, Hongping and Shi, Yilei},
  year={2026},
  eprint={2601.16771},
  archivePrefix={arXiv},
  primaryClass={cs.CV},
  url={https://arxiv.org/abs/2601.16771}
}

@misc{gao2026mirage2matterphysicallygroundedgaussian,
  title={Mirage2Matter: A Physically Grounded Gaussian World Model from Video},
  author={Gao, Zhengqing and Li, Ziwen and Wang, Xin and Huang, Jiaxin and Ren, Zhenyang and Shao, Mingkai and Zhang, Hanlue and Huang, Tianyu and Cheng, Yongkang and Guo, Yandong and Lin, Runqi and Wang, Yuanyuan and Liu, Tongliang and Zhang, Kun and Gong, Mingming},
  year={2026},
  eprint={2602.00096},
  archivePrefix={arXiv},
  primaryClass={cs.CV},
  url={https://arxiv.org/abs/2602.00096}
}

@article{guo2026huge,
  title={HUGE-Bench: A Benchmark for High-Level UAV Vision-Language-Action Tasks},
  author={Guo, Jingyu and Chen, Ziye and Li, Ziwen and Gao, Zhengqing and Huang, Jiaxin and Zhang, Hanlue and Huang, Fengming and Yao, Yu and Liu, Tongliang and Gong, Mingming},
  journal={arXiv preprint arXiv:2603.19822},
  year={2026}
}

\clearpage
\appendix
\section{Appendix}
\subsection{Brep2Text Dataset}

The Brep2Text dataset is a large-scale collection comprising 269,444 question-answer (QA) pairs, constructed based on the Text2CAD corpus. This dataset was generated by utilizing the ‘abstract' and ‘beginner' semantic levels from Text2CAD and employing Qwen-Max to formulate a corresponding question for each description. The following sections detail the generation process and statistical properties of the dataset.

\paragraph{\textbf{Data Generation with Qwen-Max.}} The foundation of our dataset relies on two semantic levels selected from the Text2CAD corpus:
\begin{itemize}
    \item \textbf{Abstract Level:} Contains high-level descriptions of the CAD model's overall geometry and topology. This level addresses the question of “what the model is".
    \item \textbf{Beginner Level:} Elaborates on the procedural steps of construction and specifies the detailed dimensions of the model. This level focuses on “how the model is built".
\end{itemize}

The core methodology involves converting the declarative descriptions within Text2CAD into ground-truth answers. For each answer, we leverage Qwen-Max to automatically generate a relevant, context-aware question. This process effectively transforms the corpus from a descriptive format into an instruction-following QA format. To guide the model's output, we engineered specific prompts for each semantic level. The prompt designed for the abstract level is detailed in Table~\ref{tab:brep2text_abstract}, while the prompt for the beginner level is shown in Table~\ref{tab:brep2text_beginner}. Representative examples of the generated data are provided in Table~\ref{tab:data_details}.

\begin{figure*}[h]
  \centering
  \includegraphics[width=0.9\linewidth]{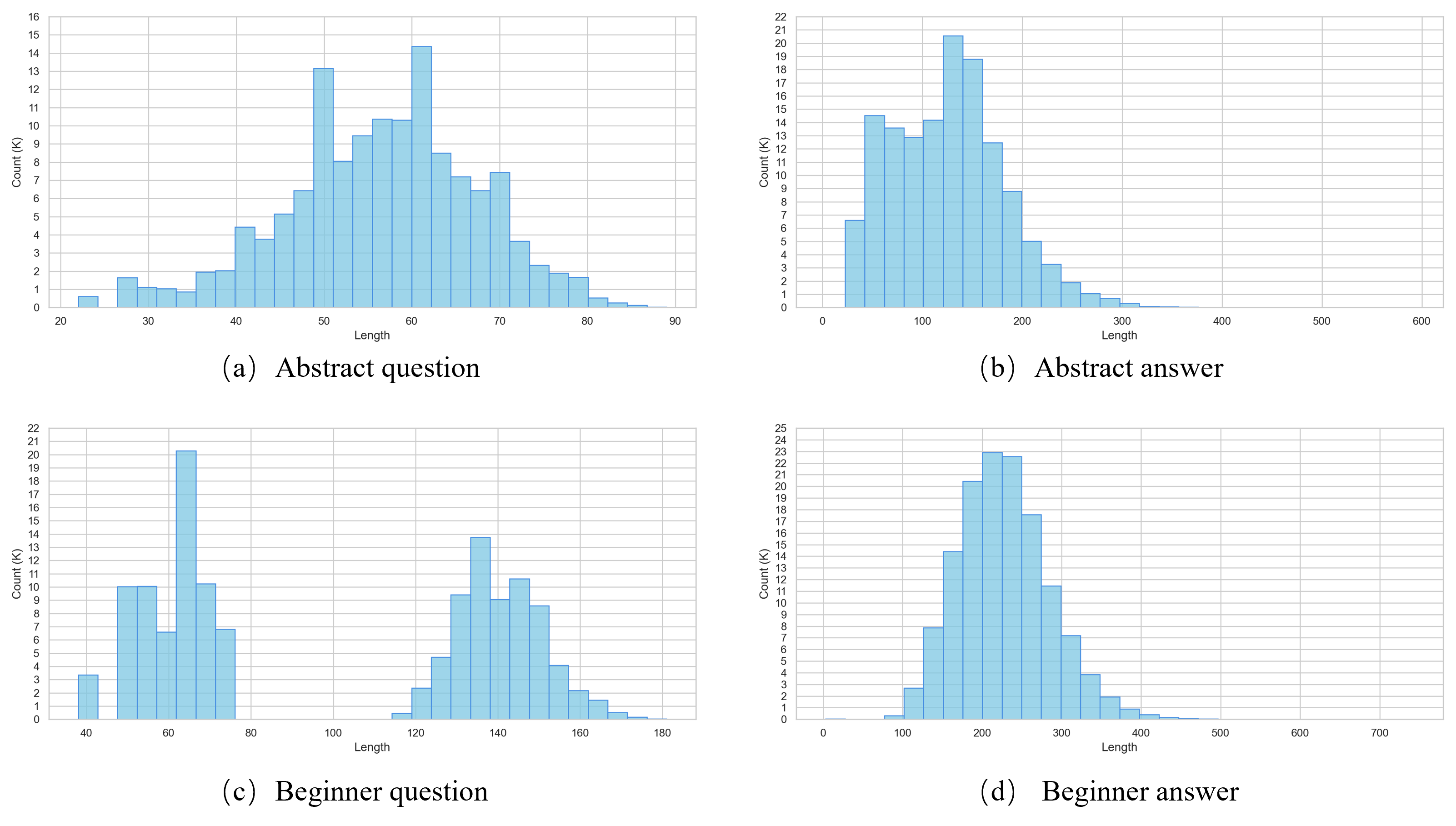}
  \caption{Length distributions in the Brep2Text dataset for (a) abstract questions, (b) abstract answers, (c) beginner questions, and (d) beginner answers.}
  \label{fig:changdu}
\end{figure*}

\begin{figure*}[h]
  \centering
  \includegraphics[width=0.9\linewidth]{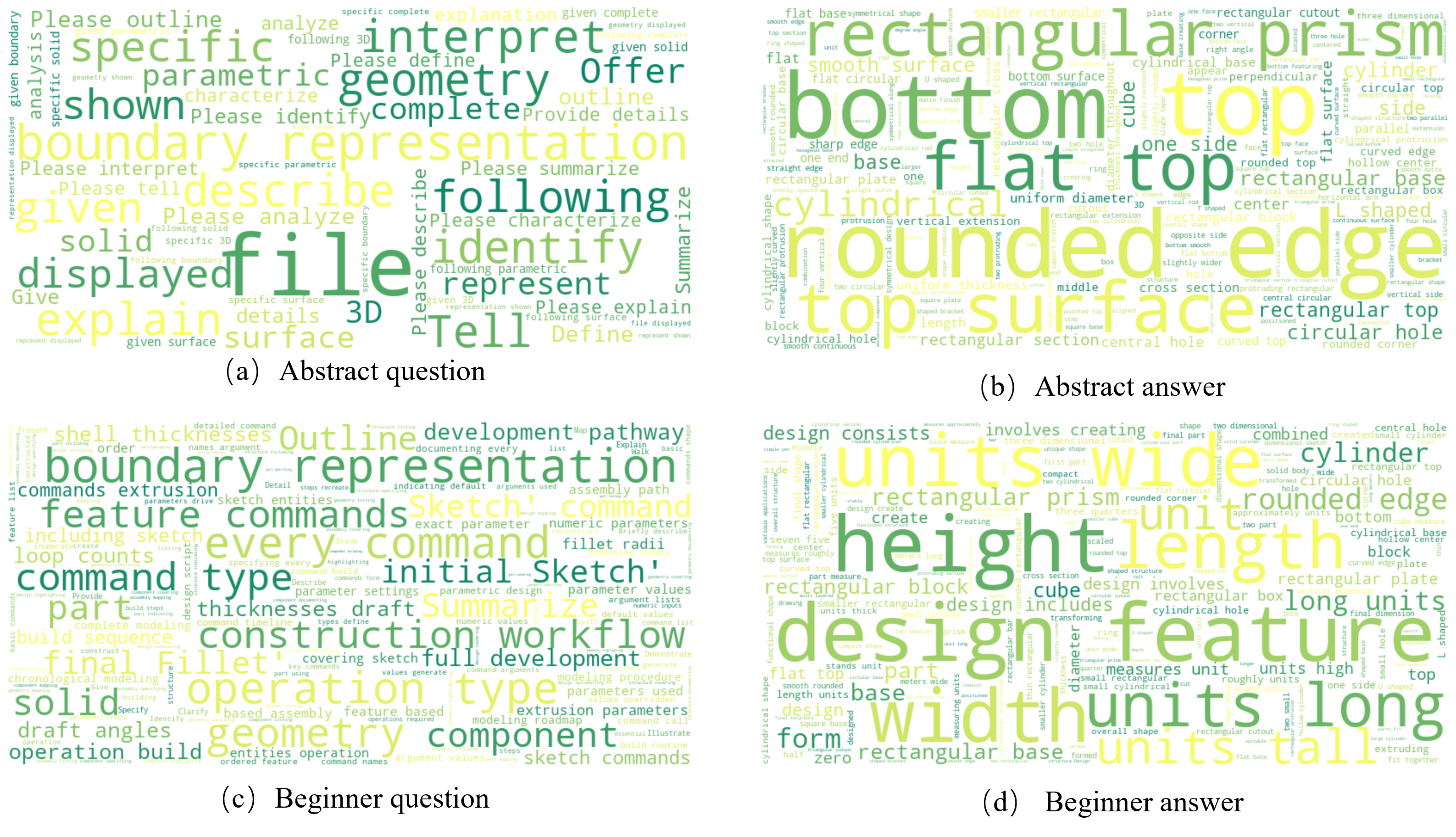}
    \caption{Word clouds for the Brep2Text dataset representing (a) abstract questions, (b) abstract answers, (c) beginner questions, and (d) beginner answers.}
    \label{fig:ciyun}
\end{figure*}

\paragraph{\textbf{Dataset Distribution Analysis.}} This section provides a statistical overview of the Brep2Text instruction-following dataset. Figure~\ref{fig:changdu} illustrates the length distributions for both questions and answers across the two semantic levels. The distributions indicate that samples from the beginner level have a greater average text length compared to those from the abstract level. In Figure~\ref{fig:ciyun}, we present word clouds generated from the dataset after the exclusion of common stop words and generic terms (e.g., “brep," “model," “cad"). The word clouds reveal distinct thematic focuses for each level. The abstract level is characterized by terms related to object identity and properties (i.e., “what it is"). In contrast, the beginner level is dominated by vocabulary associated with design actions and processes (i.e., “how to design it").

\begin{table}[h]
  \centering
  \caption{Data details, the table presents examples of questions and answers related to ‘abstract' and ‘beginner' in the dataset.}
  \label{tab:data_details}
  \begin{tabular}{p{0.95\linewidth}}
    \toprule
    \textbf{Abstract} \\
    \textbf{Question:} \textcolor{darkblue}{What} does this CAD model represent? \\
    \textbf{Answer:} A \textcolor{darkred}{rectangular plate} with \textcolor{darkred}{rounded corners} and a \textcolor{darkred}{small circular hole} at the center. \\
    \midrule
    \textbf{Beginner} \\
    \textbf{Question:} \textcolor{darkblue}{Outline} the basic CAD steps to create this Brep structure. \\
    \textbf{Answer:} Design a \textcolor{darkred}{cylinder} with a height of \textcolor{darkred}{0.75} units and a \textcolor{darkred}{square} base measuring \textcolor{darkred}{0.2586} units on each side. The cylinder is formed by \textcolor{darkred}{extruding a two-dimensional sketch} into a three-dimensional shape. \\
    \bottomrule
  \end{tabular}
  \vspace{1em}
\end{table}

\subsection{Qwen-Max Evaluation prompts}
\paragraph{\textbf{Open-vocabulary Classification.}}
In this task, Qwen3-30B is employed as an automated evaluator to determine whether the model-generated response and the description provided in the Text2CAD dataset refer to the same object type. The evaluation process is detailed in Table~\ref{tab:open_free_cls_prompt}, where \texttt{\{ground\_truth\}} represents the object description from Text2CAD, and \texttt{\{model\_output\}} is the model's generated result. This task does not require an exact textual match; a response is considered a match as long as it refers to the same object category.

\paragraph{\textbf{Object Captioning.}}
For this task, we utilize Qwen-Max to evaluate the captions generated by the model by comparing them against the captions provided in the Text2CAD dataset. Qwen-Max is instructed to identify the aspects covered in the ground truth and to calculate the proportion of these aspects that are accurately covered in the model-generated caption. The score ranges from 0 to 100, with each aspect contributing equally. The evaluation procedure is illustrated in Table~\ref{tab:object_captioning_prompt}, where \texttt{\{ground\_truth\}} denotes the caption description from Text2CAD, and \texttt{\{model\_output\}} is the model's output.

\subsection{Training Details}

All experiments are conducted on four 80 GB NVIDIA A800 GPUs using a two-stage training scheme.

\begin{itemize}
    \item \textbf{Step 1: Cross-Modal Alignment Pre-training.} 
    This stage employs the AdamW optimizer with an initial learning rate of $1 \times 10^{-4}$ and a weight decay of 0.01. A linear warmup is applied during the first five epochs, followed by a cosine annealing schedule. The model is trained for 200 epochs with a batch size of 256. Mixed-precision training and gradient clipping are adopted to ensure numerical stability during optimization.

    \item \textbf{Step 2: Two-Stage Fine-Tuning of the LLM.} 
    The fine-tuning process is divided into two stages, all using the AdamW optimizer with a weight decay of 0.05. Each stage adopts different learning rate schedules and iteration settings. In \textit{\textbf{Stage I}}, the initial learning rate is set to 3e-5, the minimum learning rate to 1e-5, and the warmup learning rate to 1e-6. Training is performed for one epoch with 70000 iterations, where the first 7000 iterations are reserved for warmup.
    In \textit{\textbf{Stage II}}, the initial learning rate is 5e-6, the minimum learning rate is 1e-6, and the warmup learning rate remains 1e-6. The model is trained for three epochs, each with 50000 iterations and 5000 warmup iterations.
\end{itemize}

\subsection{Qualitative Results}
\begin{figure*}[h]
  \centering
  \includegraphics[width=0.9\linewidth]{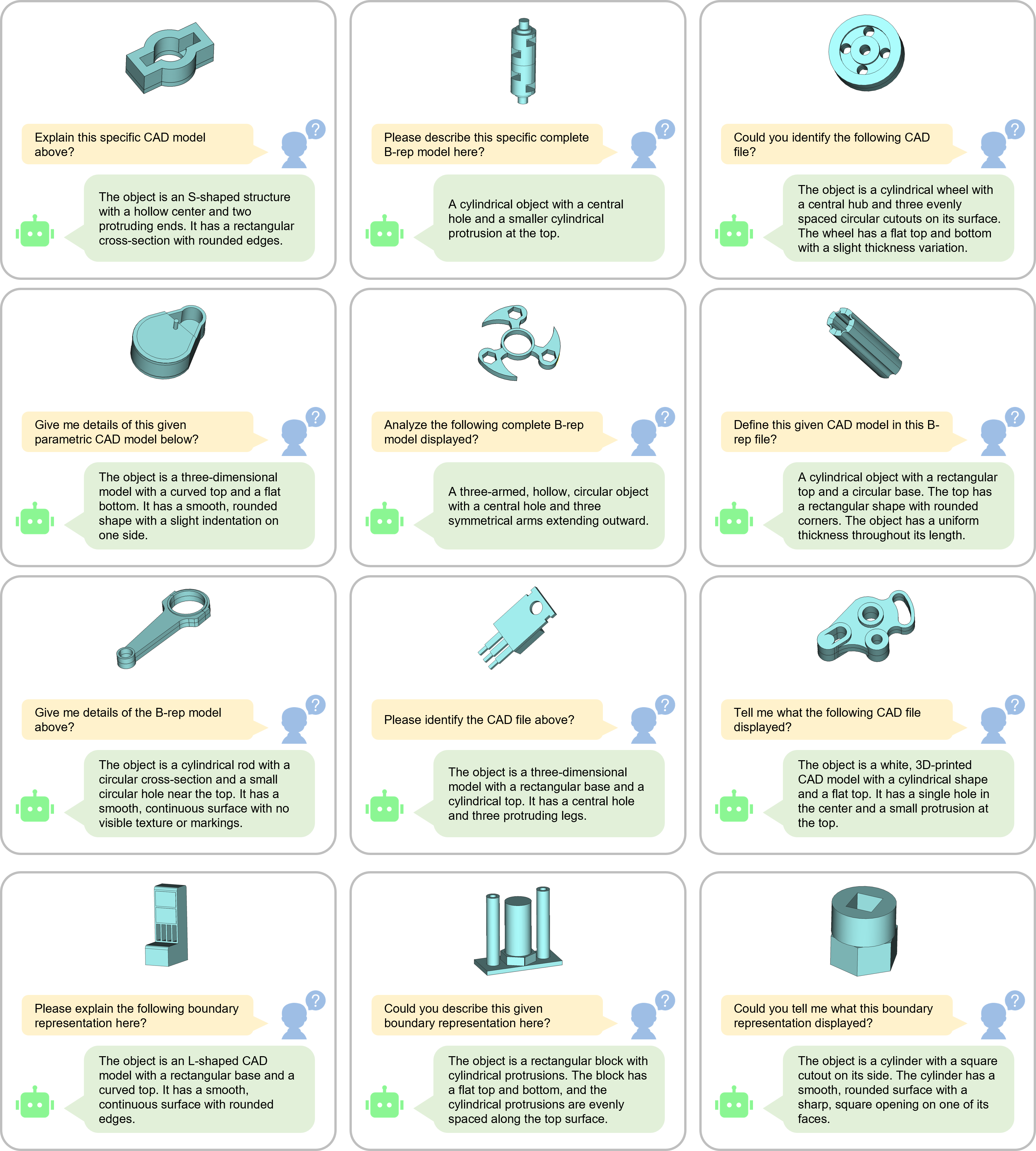}
    \caption{Dialogue examples of BrepLLM for abstract-level queries. For each given B-Rep model and user question, the model generates a corresponding response.}
    \label{fig:abstract_demo}
\end{figure*}

\begin{figure*}[h]
  \centering
  \includegraphics[width=0.9\linewidth]{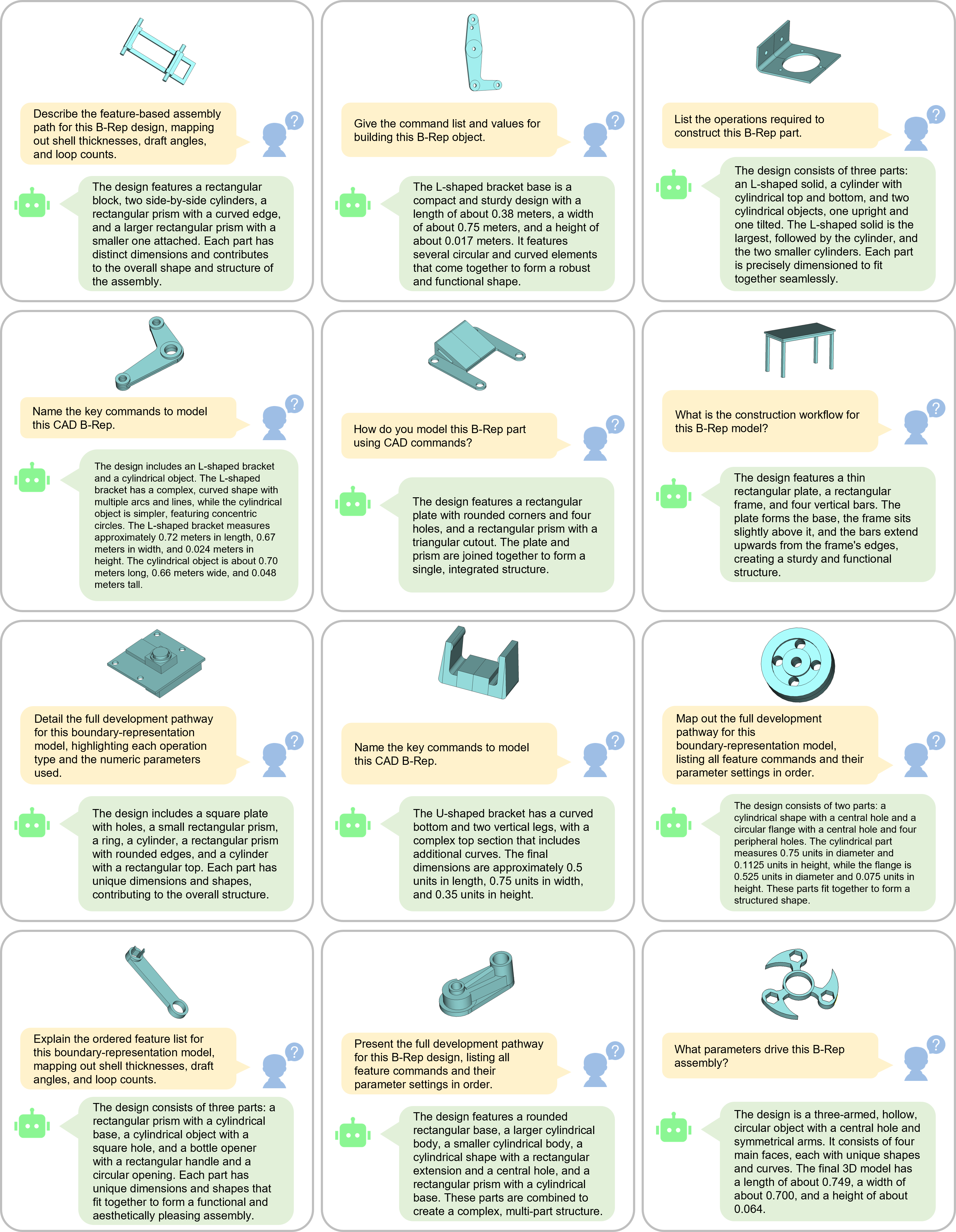}
    \caption{Dialogue examples of BrepLLM for beginner-level queries. For each given B-Rep model and user question, the model generates a corresponding response.}
    \label{fig:beginner_demo}
\end{figure*}

\begin{figure*}[h]
  \centering
  \includegraphics[width=0.9\linewidth]{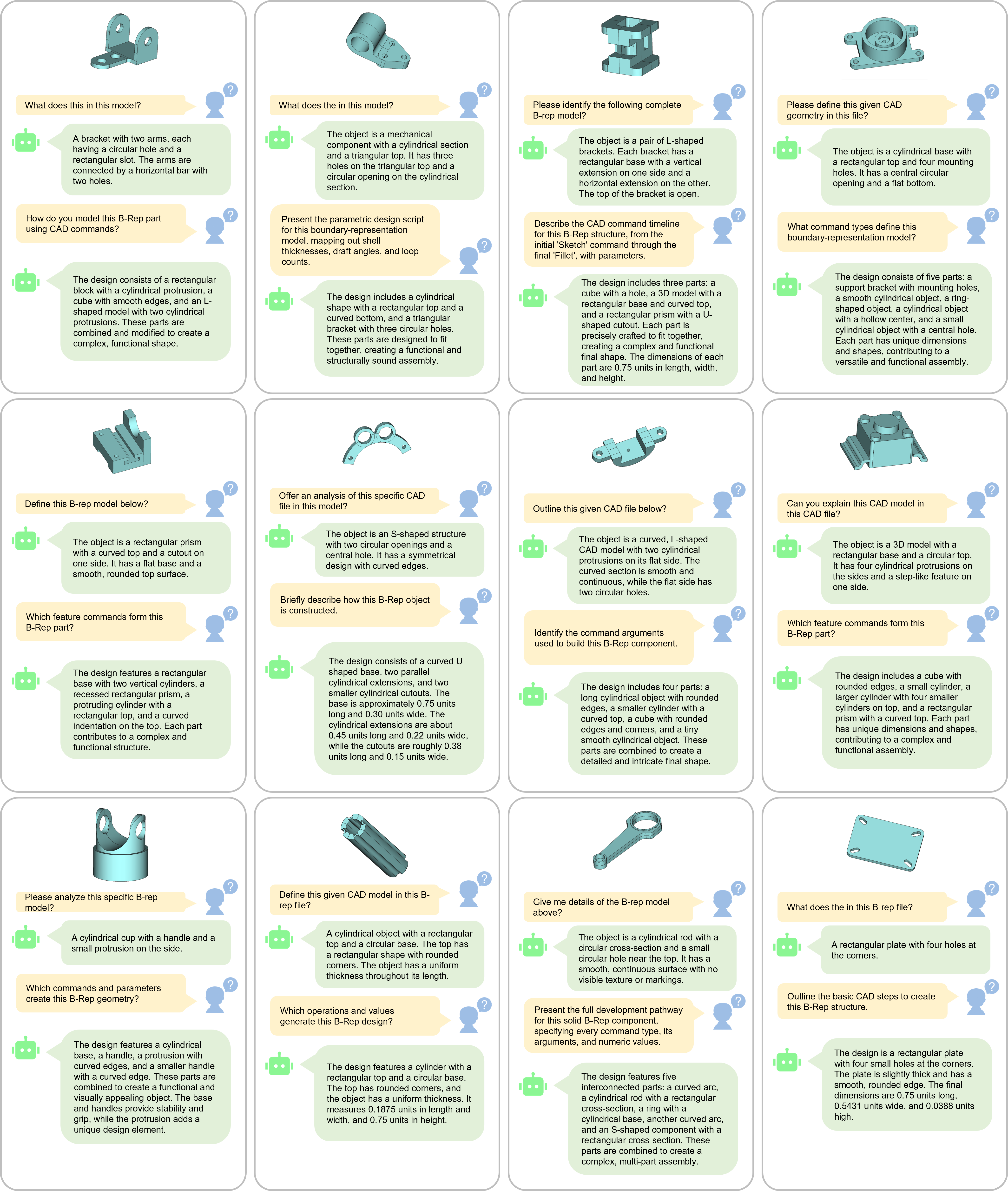}
    \caption{Examples of interactive sessions between BrepLLM and a human user.}
    \label{fig:conversation_demo}
\end{figure*}

\paragraph{\textbf{Response Capability of BrepLLM.}}
The qualitative response capabilities of BrepLLM were evaluated on two distinct semantic levels: \textit{\textbf{abstract}} and \textit{\textbf{beginner}}. On the abstract level, as shown in Figure~\ref{fig:abstract_demo}, BrepLLM accurately identifies the overall shape characteristics of B-Rep objects. On the beginner level, illustrated in Figure~\ref{fig:beginner_demo}, the model demonstrates a fine-grained understanding of the construction process and dimensional information contained within the B-Rep files. As further evidenced by the qualitative comparisons under identical samples and prompts in Table~\ref{tab:qual_compare}, BrepLLM produces responses that are consistently more accurate and comprehensive than competing models. For instance, in Sample~1, BrepLLM correctly infers salient geometric details such as the slanted side faces, the symmetrically placed hollow triangular braces, and the centered rectangular notch along the bottom edge.
\paragraph{\textbf{Multi-turn Dialogue.}} Figure~\ref{fig:conversation_demo} presents an example of multi-turn interaction between BrepLLM and a human user, highlighting its ability to comprehend the shape, appearance, and topological features of B-Rep data. Notably, BrepLLM remains robust under occlusion and is capable of identifying geometric properties such as whether a sphere is hollow—tasks that are challenging for vision-based inputs alone. These results indicate that BrepLLM exhibits strong capability in understanding B-Rep representations and in effectively responding to user instructions.

\begin{table}[!t]
\centering
\renewcommand{\arraystretch}{1.3}

\caption{An example of the data generation process for an \textbf{abstract-level} sample in the Brep2Text dataset using Qwen-Max. The table illustrates how a declarative description from Text2CAD (the \textbf{Input}) is treated as a ground-truth answer. Guided by the \textbf{System Prompt}, Qwen-Max automatically formulates a corresponding question (the \textbf{Output}).}
\label{tab:brep2text_abstract}

\resizebox{\textwidth}{!}{
\begin{tabular}{p{3cm} p{13.5cm}} 
\toprule
& \includegraphics[width=0.3\linewidth]{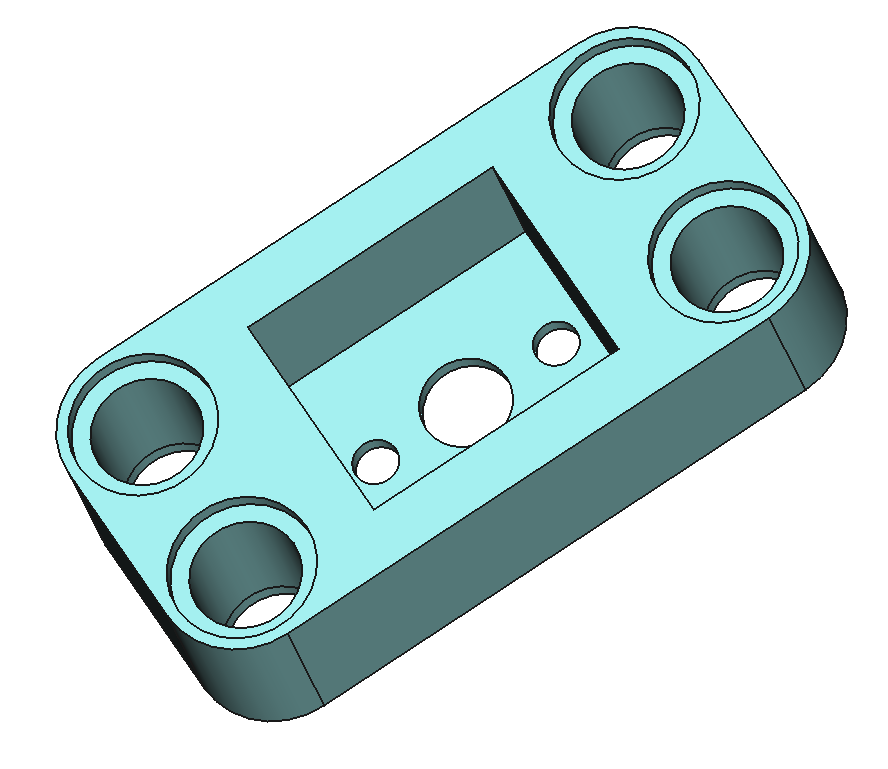} \\
\midrule
\textbf{uid} & 0009/00098309.step \\
\midrule
\textbf{System Prompt} & 
You are a data generator for a CAD Q\&A dataset. \\
& Given a short semantic description of a CAD model, write exactly ONE concise interrogative sentence that would be naturally answered by that description. \\
& Constraints: \\
& Output exactly one question, one sentence. \\
& Do Not add any extra text, labels, or quotes. \\
& \\
& Examples: \\
& ANSWER: A rectangular plate with rounded corners and a small circular hole at the center. \\
& QUESTION: What does this CAD model represent? \\
\midrule
\textbf{Input} & The object is a rectangular bracket with a central rectangular cutout and four cylindrical holes on its sides. \\
\midrule
\textbf{Output} & What does this CAD model represent in this model? \\
\bottomrule
\end{tabular}
}
\end{table}

\begin{table*}[!t]
\centering
\renewcommand{\arraystretch}{1.3}
\caption{An example of the data generation process for an \textbf{beginner-level} sample in the Brep2Text dataset using Qwen-Max. The table illustrates how a declarative description from Text2CAD (the \textbf{Input}) is treated as a ground-truth answer. Guided by the \textbf{System Prompt}, Qwen-Max automatically formulates a corresponding question (the \textbf{Output}).}
\label{tab:brep2text_beginner}
\resizebox{\textwidth}{!}{
\begin{tabular}{p{3cm} p{13.5cm}}
\toprule
& \includegraphics[width=0.3\linewidth]{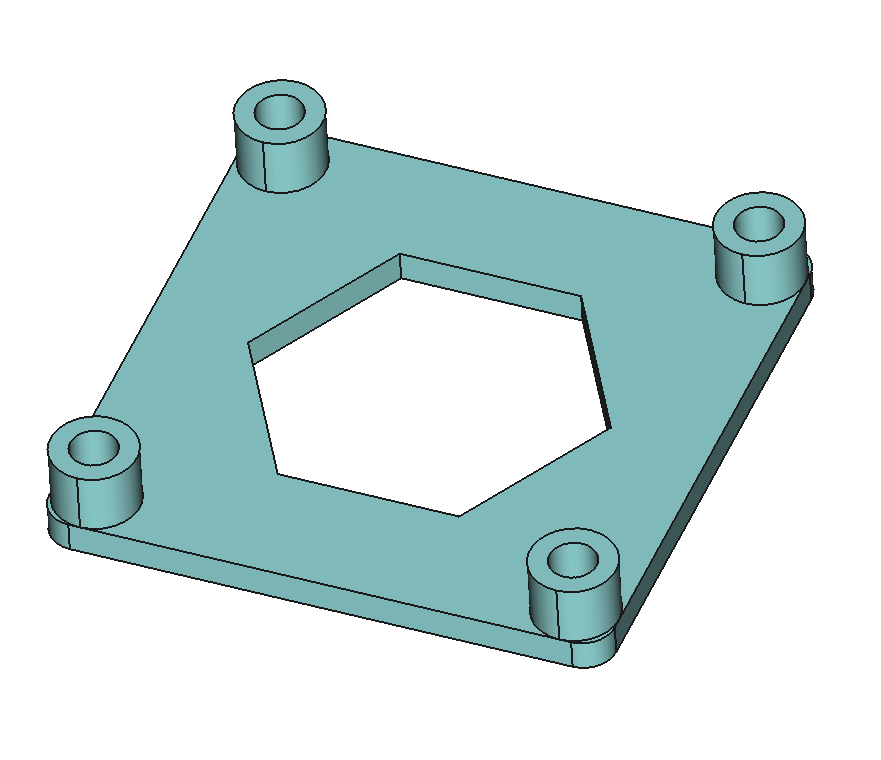} \\
\midrule
\textbf{uid} & 0010/00108226.step \\
\midrule
\textbf{System Prompt} & 
You are a data generator for a CAD Q\&A dataset. Given the answer, provide a question about how to construct this CAD model. \\
& Constraints: \\
& Output exactly one question, one sentence. \\
& Do NOT add any extra text, labels, or quotes. \\
& \\
& Examples: \\
& ANSWER: Design a cylinder with a height of 0.75 units and a square base measuring 0.2586 units on each side. The cylinder is formed by extruding a two-dimensional sketch into a three-dimensional shape. \\
& QUESTION: Outline the basic CAD steps to create this B-Rep structure. \\
\midrule
\textbf{Input} & The design features a square plate with a hexagonal hole, four smaller cylinders arranged on top of the plate, and four taller cylinders with rounded ends positioned above the smaller ones. The overall structure is multi-layered and symmetrical. \\
\midrule
\textbf{Output} & Specify the ordered feature list for this CAD B‑Rep part, highlighting each operation type and the numeric parameters used. \\
\bottomrule
\end{tabular}
}
\end{table*}

\newcolumntype{L}[1]{>{\raggedright\arraybackslash}p{#1}}
\newcolumntype{Y}{>{\raggedright\arraybackslash}X}

\begin{table}[t]
\centering
\caption{The prompt and examples for the open-vocabulary classification task. Qwen-Max is prompted to act as an automated evaluator, determining if a model-generated output and the ground-truth description refer to the same object category.}
\label{tab:open_free_cls_prompt}

\scriptsize                         
\setlength{\tabcolsep}{3pt}         
\renewcommand{\arraystretch}{1.08}  

\begin{tabularx}{\columnwidth}{@{}L{0.26\columnwidth}Y@{}}
\toprule
\textbf{Prompt Task} & Open-vocabulary Classification \\
\midrule
\textbf{Model} & Qwen-Max \\
\midrule
\textbf{Prompt} &
You will analyze two short descriptions about CAD models or designs. Decide if they refer to the SAME CAD object/feature CATEGORY based on geometric/topological intent and primitive class (e.g., washer/annulus, cylinder/rod, cuboid/block, bracket, gear, torus, hinge; features like through-hole, slot, pocket). Ignore all numeric dimensions, units, materials, finishes, and style words. Treat close synonyms as the same category (e.g., ring/washer/annulus; rod/cylinder), unless the geometric class truly changes (e.g., flat annular washer vs solid torus are different). \par\medskip
Now, analyze the following:\par
\textbf{Input}: 1.\texttt{\{ground\_truth\}} 2.\texttt{\{model\_output\}}\par
\textbf{Output}: \\
\midrule

\textbf{Example 1} &
\textbf{Input}: 1. A ring-shaped object with a hollow center and flat faces (washer). 2. design a flat, circular ring with a central hole; 0.75 wide, 0.075 thick. \par
\textbf{Output}: T\#Both describe a flat annular washer; dimensions ignored. \\
\midrule

\textbf{Example 2} &
\textbf{Input}: 1. Cylindrical peg. 2. A metal rod made by extruding a circle. \par
\textbf{Output}: T\#Both are cylinders/rods. \\
\bottomrule
\end{tabularx}
\end{table}

\begin{table*}[!t]
\centering
\caption{The prompt and example for the object captioning evaluation task. Qwen-Max is prompted to act as an automated evaluator, scoring the alignment between a model-generated caption and the ground-truth caption on a scale of 0-100.}
\label{tab:object_captioning_prompt}

\scriptsize
\setlength{\tabcolsep}{3pt}
\renewcommand{\arraystretch}{1.08}

\begin{tabularx}{\textwidth}{@{}>{\raggedright\arraybackslash}p{0.26\textwidth} >{\raggedright\arraybackslash}X@{}}
\toprule
\textbf{Prompt Task} & Object Captioning Evaluation \\
\midrule
\textbf{Model} & Qwen-Max \\
\midrule
\textbf{Prompt} &
Please evaluate the overall match between the Model Caption and the Ground Truth Caption for a CAD model. The model should extract aspects to be evaluated from the Ground Truth Caption and assign equal weight to each aspect (e.g., primitive/category, key features, topology/structure). Partial matches of synonyms/family concepts are allowed (e.g., ring $\approx$ washer $\approx$ torus, cylinder $\approx$ rod/pin, rectangular block $\approx$ plate/beam). Values are only scored if both sides mention the same dimension name (length/width/height/radius/diameter/thickness, etc.) and are assumed to use the same units. An absolute error $\le$ 0.2 is considered correct. If the ground truth contains a value but the model does not, that dimension is not scored. If the model includes a value not mentioned in the ground truth, it is ignored — neither rewarded nor penalized. Fabricated features that contradict the ground truth will result in point deductions. The score ranges from 0 to 100, with each aspect contributing equally. Partial scores can be given for similar concepts. \par\medskip
Now score the following: \par
\textbf{Ground Truth}: \texttt{\{ground\_truth\}} \par
\textbf{Model}: \texttt{\{model\_output\}} \par
\textbf{Output:} \\
\midrule

\textbf{Example} &
\textbf{Ground Truth:} A small rectangular block; length 0.75, width 0.375, height 0.204. \par
\textbf{Model:} rectangular block; length 0.75, width 0.375, height 0.375. \par
\textbf{Output:} 100\#primitive match; dimensions within 0.2 tolerance. \\
\bottomrule
\end{tabularx}
\end{table*}

\begin{table}[!t]
\centering
\caption{Qualitative comparison for 3D object captioning on Brep2Text dataset against the next-best baseline. Green highlights indicate salient details that align with the ground truth.}
\label{tab:qual_compare}
\scriptsize
\setlength{\tabcolsep}{3pt}
\renewcommand{\arraystretch}{1.08}

\begin{tabular}{>{\raggedright\arraybackslash}p{0.16\linewidth}
                >{\raggedright\arraybackslash}p{0.39\linewidth}
                >{\raggedright\arraybackslash}p{0.39\linewidth}}
\toprule
\textbf{Samples~1,~2}
&
\multicolumn{1}{c}{\includegraphics[height=1.25cm]{images/case1.png}} &
\multicolumn{1}{c}{\includegraphics[height=1.25cm]{images/case2.png}} \\
\midrule

\textbf{Prompt} &
\emph{What is this?} &
\emph{How this CAD model was constructed, describe this CAD model in detail.} \\

\textbf{Ground Truth} &
This is a rectangular plate with two triangular braces on its front face and a flat back; both side edges are beveled and slope upward, and there is a centered rectangular notch along the bottom edge. &
The design features a rounded rectangular base, a larger cylindrical body, a smaller cylindrical body, a cylindrical shape with a rectangular extension and a central hole, and a rectangular prism with a cylindrical base. \\
\midrule

Qwen3-VL-8B &
This is a rectangular plate-like part with two symmetrical hollow triangular support structures on the front face. Overall, it is relatively flat and could serve as a bracket or connector. There seems to be a rectangular opening in the middle of the bottom edge, and the two side edges appear to be chamfered.
&
The model is based on a rounded-rectangle base, formed by extrusion to create the main body. Cylindrical bosses are added at both ends of the base (one larger and the other smaller), and holes are cut through the centers of both bosses. The two bosses are connected by a combination of a prism and a triangular component. \\
\midrule

PointLLM-13B &
This is a rectangular plate-type CAD part with two symmetrical triangular protrusions on the front face. There is a rectangular notch in the middle of the bottom edge, and overall it looks like a mounting part for positioning or snap-fit attachment.
&
This part is obtained by extruding a rounded-rectangle base. There is a circular protrusion on each side of the base. The two cylinders are connected by a rectangular block. \\
\midrule

ShapeLLM-13B &
This is a flat rectangular component with two symmetrical triangular ribs on the front surface to enhance stiffness; the back is likely flat. The side edges are beveled, forming a slanted transition along the edges.
&
This CAD model appears to be created by first generating a rounded-rectangle base plate, then adding some hollow cylinders on top. There is a hollow cylindrical boss at each end, and the middle section is connected by a prism. \\
\midrule

MiniGPT-3D &
This is a rectangular CAD model with two triangular brackets on each side, integrated into the rectangular plate plane and with a cut at the bottom.
&
This model consists of a rectangular base, a large cylinder, and a small cylinder. It is made by stretching a rectangular base and adding circular features on both sides. \\
\midrule

\textbf{BrepLLM} &
\textbf{This is a CAD model based on a \textcolor{green!50!black}{flat rectangular plate}. It features a centered \textcolor{green!50!black}{rectangular notch} on the bottom edge and a flat back. The front face has two \textcolor{green!50!black}{symmetrically positioned} triangular braces \textcolor{green!50!black}{with internal triangular cutouts}, and both side edges are slanted faces.}
&
\textbf{The model is built on a \textcolor{green!50!black}{rectangular base with rounded corners}. It contains \textcolor{green!50!black}{a larger cylinder} and \textcolor{green!50!black}{a smaller cylinder} at \textcolor{green!50!black}{each end}. It combines a \textcolor{green!50!black}{cylindrical body} with a rectangular extension and a \textcolor{green!50!black}{central hole}, as well as a straight prism with a \textcolor{green!50!black}{cylindrical base}.} \\
\bottomrule
\end{tabular}
\end{table}

\end{document}